%% file: main.tex
\documentclass{article}

 \usepackage[main, final]{neurips_2026}

\usepackage[utf8]{inputenc} 
\usepackage[T1]{fontenc}    
\usepackage{hyperref}       
\usepackage{url}            
\usepackage{booktabs}       
\usepackage{amsfonts}       
\usepackage{nicefrac}       
\usepackage{microtype}      
\usepackage{xcolor}         

\usepackage{amsmath}
\usepackage{booktabs}
\usepackage{multirow}
\usepackage{makecell}
\usepackage{graphicx}
\usepackage{cleveref}
\usepackage{bm}
\usepackage{hyperref}
\usepackage{wrapfig}
\usepackage{svg}
\usepackage{caption}
\usepackage[ruled]{algorithm2e}

\newcommand{\MethodName}{D\textsuperscript{3}S\textsuperscript{2}}

\title{\MethodName: Diffusion-Guided Dataset Distillation for Semantic Segmentation}

\author{
Wenjie Zheng
,
Haoji Hu\thanks{Corresponding author.}
,
Jiali Lu
,
Xingze Zou
,
Jing Wang \\
Zhejiang University \\
\texttt{\{wenjie\_zheng, haoji\_hu\}@zju.edu.cn} 
}

\begin{document}

\maketitle

\begin{abstract}
Dataset distillation (DD) aims to compress large-scale datasets into compact synthetic sets while preserving training efficacy. However, existing studies mainly focus on image classification, leaving dense prediction tasks such as semantic segmentation largely underexplored. In this work, we identify three key challenges for segmentation DD: (\textit{i}) long-tailed class imbalance, (\textit{ii}) the need for strict pixel-wise alignment between images and dense labels, and (\textit{iii}) the high computational cost of optimizing high-resolution data with complex models. To address these challenges, we propose \textbf{\MethodName}, a \textbf{D}iffusion-guided \textbf{D}ataset \textbf{D}istillation framework for \textbf{S}emantic \textbf{S}egmentation. Our method adopts a two-stage design. In \textbf{Class-Balanced Mask Selection}, we construct a representative mask set via a greedy strategy that prioritizes underrepresented classes. In \textbf{Diffusion-Guided Image Synthesis}, we employ a pretrained layout-to-image diffusion model to generate images conditioned on the selected masks, naturally ensuring spatial alignment. To further enhance the training utility of synthesized data, we introduce guided diffusion sampling with two complementary objectives: a segmentation-consistency loss for pixel-level alignment, and a class-wise feature matching loss for aligning per-class feature statistics across layers. Extensive experiments demonstrate the superiority of \MethodName. Notably, at an extremely compression rate of 1\%, our method achieves 24.99\% and 35.49\% mIoU on ADE20K and COCO-Stuff with Mask2Former (Swin-S), outperforming random selection by 9.34\% and 5.70\%, respectively. Our code is available at \url{https://github.com/zwj084/D3S2}.
\end{abstract}

\section{Introduction}
\begin{figure}[!htbp]
    \centering
    \includegraphics[width=\textwidth]{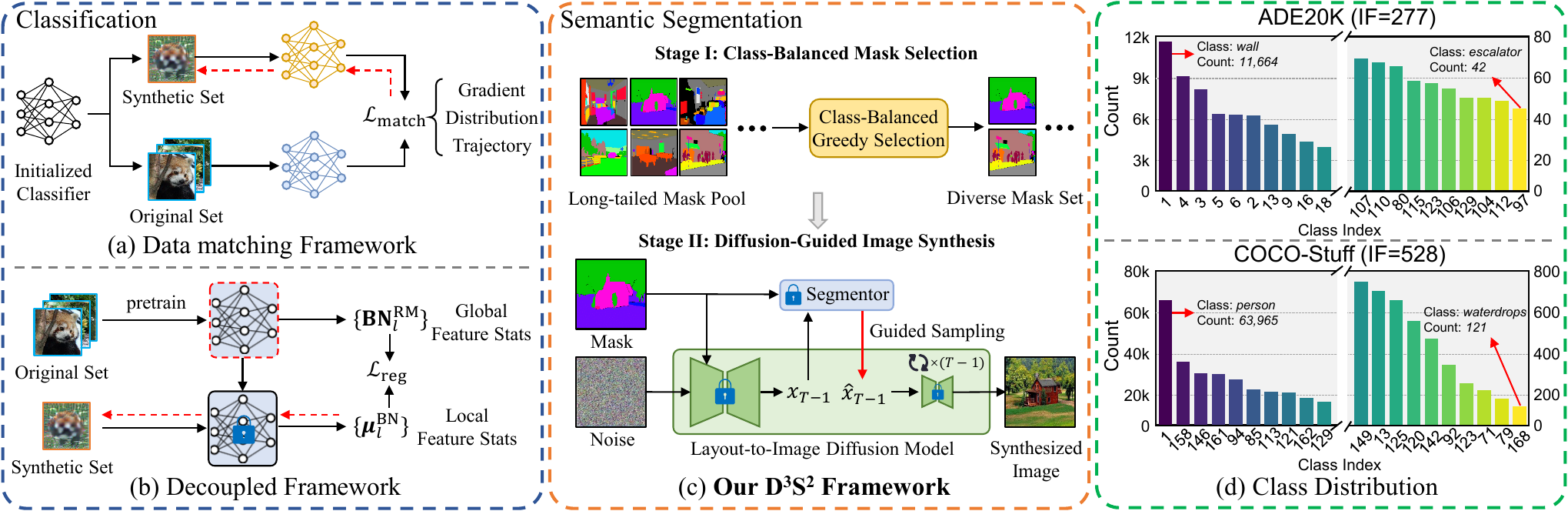}
    \caption{(a) Data-matching methods; (b) Decoupled methods; (c) Our proposed \MethodName, the first DD framework tailored for semantic segmentation, which generates distilled data via a layout-to-image diffusion model; (d) Class distributions of ADE20K~\cite{ade20k} and COCO-Stuff~\cite{coco_stuff}.}
    \vspace{-4mm}
    \label{fig:teaser}
\end{figure}

Dataset Distillation (DD)~\cite{Dataset_distillation,DD_survey_1,DD_survey_2} aims to compress large-scale datasets into compact synthetic sets while preserving training efficacy. By reducing storage and computational costs, DD has emerged as a promising paradigm for efficient model training under resource constraints. Recent advances have achieved strong performance in image classification via data matching techniques (see \Cref{fig:teaser} (a)), including Gradient Matching~\cite{GM_1,GM_2,GM_3}, Distribution Matching~\cite{DM_1,DM_2,DM_3}, and Training Trajectory Matching~\cite{TTM_1,TTM_2}. More recently, decoupled methods~\cite{SRe2L,Decoupled_2,Decoupled_3} (see \Cref{fig:teaser} (b)) disentangle model training from data optimization, avoiding costly bi-level optimization and improving scalability.

Despite these advances, DD remains largely limited to classification, leaving dense prediction tasks such as semantic segmentation underexplored. Extending DD to segmentation is non-trivial due to three key challenges: (\textit{i}) \textbf{Long-tailed Class Imbalance.} Unlike relatively balanced classification datasets~\cite{imagenet,cifar}, segmentation datasets~\cite{ade20k,coco_stuff} exhibit highly skewed pixel distributions (\Cref{fig:teaser} (d)), making it difficult to allocate a limited distillation budget effectively.
(\textit{ii}) \textbf{Dense Label Coupling with Locality Requirement.} In contrast to classification-based DD methods that operate on global image-level supervision, existing paradigms (e.g., SRe2L~\cite{SRe2L}) mainly rely on aligning global feature statistics such as BatchNorm statistics, which inherently discard spatial structure. However, semantic segmentation requires pixel-wise supervision, where object semantics are tightly coupled with spatial layouts. This introduces a fundamental gap: DD methods designed for classification optimize images at a global representation level, while segmentation demands fine-grained, region-aware consistency between synthesized images and masks, making purely global feature alignment insufficient.
(\textit{iii}) \textbf{High Computational Overhead.} The increased complexity of segmentation models and high-resolution supervision makes per-sample optimization (typically 1k--4k iterations) prohibitively expensive, limiting the direct applicability of existing DD paradigms.

To address these challenges, we propose \textbf{\MethodName}, a \textbf{D}iffusion-guided \textbf{D}ataset \textbf{D}istillation framework tailored for \textbf{S}emantic \textbf{S}egmentation. As shown in \Cref{fig:teaser} (c), our method follows a two-stage design. \textbf{Stage I: Class-Balanced Mask Selection.} To combat long-tailed class imbalance, we construct a compact yet representative mask set. Specifically, we introduce a class-balanced greedy selection strategy that iteratively selects samples using a scoring function prioritizing underrepresented classes, promoting class diversity and coverage under a limited distillation budget.
\textbf{Stage II: Diffusion-Guided Image Synthesis.} To enforce strict pixel-wise alignment while avoiding prohibitive per-sample optimization, we build on recent advances in Generative Dataset Distillation~\cite{Glad,Dim,Minimax_diffusion,D4M,IGD,LD3M,LGD} and exploit the strong generative priors of diffusion models~\cite{SDE,DDPM,DDIM,LDM}. We employ a layout-to-image diffusion model~\cite{freestyle} pretrained on the target dataset to synthesize images conditioned on selected masks, naturally enforcing pixel-level alignment. To further improve the training utility of the synthesized data, we introduce guided diffusion sampling~\cite{UGD,Freedom} with two complementary objectives. A segmentation-consistency loss enforces pixel-level alignment with target masks via a pretrained segmentation model, while a class-wise feature matching loss aligns per-class feature statistics across layers. Unlike conventional global BatchNorm statistics matching~\cite{SRe2L}, we compute per-class feature statistics from the full dataset and align them with those extracted from the generated samples. This design captures class-specific characteristics and promotes localized feature refinement. Notably, all guidance objectives are integrated directly into the diffusion sampling process. This enables highly time-efficient generation of high-quality distilled data within a small number of sampling steps, requiring less than one minute to synthesize a single image.

We conduct extensive experiments on ADE20K~\cite{ade20k} and COCO-Stuff~\cite{coco_stuff}. Across a wide range of distillation ratios, \MethodName\ consistently achieves top-tier performance on multiple segmentation architectures, including Mask2Former~\cite{mask2former} and SegFormer~\cite{Segformer} with various backbones, demonstrating strong robustness and generalization. For instance, the full COCO-Stuff training set contains 118,287 images, and training Mask2Former (Swin-S) typically requires about two days to reach 48.19\% mIoU. In contrast, with only 0.25\% data (295 distilled images), \MethodName\ finishes training in roughly 3 hours and achieves 26.87\% mIoU, outperforming random selection by 7.99\%. Although a performance gap remains compared to full-data training, reflecting the challenge of DD for semantic segmentation, these results suggest that diffusion-guided DD for semantic segmentation is a promising direction, with substantial room for improvement. Our main contributions are summarized as follows:
\vspace{-2mm}
\begin{itemize}
    \item To the best of our knowledge, we propose \textbf{\MethodName}, \textbf{the first dataset distillation framework tailored for semantic segmentation.} It adopts a two-stage design, where \textit{Class-Balanced Mask Selection} constructs a representative mask set, and \textit{Diffusion-Guided Image Synthesis} efficiently generates high-quality distilled data via guided diffusion sampling.

    \item We introduce two complementary guidance objectives within the diffusion sampling process: a segmentation-consistency loss for pixel-level alignment and a class-wise feature matching loss for class-specific feature fidelity through localized statistics alignment.

    \item Extensive experiments demonstrate the superiority of \MethodName. Notably, at a 1\% compression ratio, it achieves 24.99\% and 35.49\% mIoU on ADE20K and COCO-Stuff, respectively, using Mask2Former (Swin-S), surpassing random selection by 9.34\% and 5.70\%.
\end{itemize}

\section{Related Works}

\textbf{Dataset Distillation.} Early DD methods formulate the problem as a meta-learning task~\cite{Dataset_distillation,KIP,RFAD,FRePo}, optimizing a compact synthetic set such that models trained on it generalize well to the original dataset. Subsequent works move toward a more scalable data matching paradigm, aligning training dynamics between synthetic and real data. Representative approaches include Gradient Matching~\cite{GM_1,GM_2,GM_3}, Distribution Matching~\cite{DM_1,DM_2,DM_3}, and Training Trajectory Matching~\cite{TTM_1,TTM_2}, which respectively match gradients, feature distributions, and long-range optimization trajectories.

\textbf{Decoupled Methods.} Decoupled approaches improve scalability by separating model training from data optimization. SRe2L~\cite{SRe2L} proposes a Squeeze-Recover-Relabel pipeline, which first squeezes dataset statistics into a model, then recovers synthetic data via BatchNorm statistics matching, and finally refines them with soft labels. Follow-up works~\cite{Decoupled_2,RDED,EDC} extend statistical matching to broader architectures and further enhance sample quality and diversity. Despite their success, these methods are mainly designed for image classification. Inspired by SRe2L, we argue that dense pixel-wise information in semantic segmentation can also be encoded into a segmentor and subsequently recovered. However, extending decoupled methods to segmentation is non-trivial. First, pixel-wise optimization becomes computationally expensive for high-resolution dense prediction. Second, BN statistics rely on globally averaged features and thus fail to capture class-specific semantics. While prior work such as LPLD~\cite{LPLD} has explored class-wise statistical matching for classification, applying this to segmentation requires fundamentally different treatment: in classification, class-wise statistics are computed from single-class images with well-defined support; in segmentation, multi-class scenes with variable region sizes demand careful handling to avoid conflating genuine appearance variation with spatial artifacts. To address these issues, we replace explicit pixel optimization with guided diffusion sampling~\cite{UGD,Freedom}, enabling efficient data recovery while preserving pixel-level alignment with masks. In addition, we introduce a class-wise feature matching objective that aligns per-class statistics, providing more discriminative supervision than global BN matching.

\textbf{Diffusion-based Dataset Distillation.} Recent works~\cite{Minimax_diffusion,D4M,IGD,MGD3,CaO2,DDVLCP,LD3M,DAP,LGD} leverage diffusion models to synthesize realistic and representative datasets, alleviating the limitations of direct pixel-space optimization. Several methods incorporate guided diffusion sampling to improve data quality and training effectiveness. For example, IGD~\cite{IGD} adopts influence-guided sampling, MGD$^3$~\cite{MGD3} introduces mode guidance to enhance diversity, and LGD~\cite{LGD} employs learnability-aware guidance to generate complementary samples. Despite their promising performance, these methods are primarily designed for image classification and do not account for the structured pixel-level supervision required by semantic segmentation.

\textbf{Layout-to-Image Diffusion.} Diffusion models~\cite{SDE,DDPM,DDIM,LDM} have achieved remarkable success in image synthesis and can be extended to conditional generation with structural guidance such as semantic layouts. ControlNet~\cite{ControlNet} introduces additional branches for condition injection, while T2I-Adapter~\cite{T2i-adapter} employs lightweight adapters with minimal architectural changes. In this work, we adopt the Freestyle~\cite{freestyle}, which enables precise layout-conditioned generation without additional modules, making it a lightweight and efficient choice for our framework.

\section{Preliminaries}

\textbf{Dataset Distillation for Semantic Segmentation.} DD aims to compress a large real dataset $\mathcal{T}$ into a compact synthetic dataset $\mathcal{S}$. We extend the standard DD formulation to semantic segmentation, where each image is associated with dense, pixel-wise annotations. The real dataset $\mathcal{T}$ with $N$ labeled images is defined as $\mathcal{T}=\{(\bm{x}_i,\bm{m}_i)\}_{i=1}^N$, where $\bm{x}_i \in \mathbb{R}^{H_i \times W_i \times 3}$ is the input image and $\bm{m}_i \in \mathbb{R}^{H_i \times W_i}$ is the corresponding segmentation mask, with $\bm{m}_{i}(u,v) \in \{0,1,\dots,K-1\}$ and $K$ denoting the number the classes. The distilled dataset $\mathcal{S}$ is formulated as $\mathcal{S}=\{(\hat{\bm{x}}_j,\hat{\bm{m}}_j)\}_{j=1}^{\vert \texttt{IPD} \vert}$, where $\hat{\bm{x}}_j$ and $\hat{\bm{m}}_j$ denote the synthesized images and masks, respectively, and $\vert \texttt{IPD} \vert \ll N $ represents \textit{images per dataset}, reflecting the compression ratio~\cite{OD3}. The goal of DD is to ensure that a model with weights $\theta_{\mathcal{S}}$ trained on $\mathcal{S}$ achieves performance comparable to a model trained on $\mathcal{T}$ with weights $\theta_{\mathcal{T}}$. Formally, the performance gap is bounded by:
\begin{equation}
    sup\{\vert \mathcal{L}_{\theta_{\mathcal{T}}} - \mathcal{L}_{\theta_{\mathcal{S}}} \vert \}_{(\bm{x}_v,\bm{m}_v)\in \mathcal{T}^{\prime}} \le \epsilon_{DD},
\end{equation}
where $\mathcal{L}$ denotes the loss function, and $(\bm{x}_v,\bm{m}_v)\in \mathcal{T}^{\prime}$ is some test or val set associated with $\mathcal{T}$.

\textbf{Guided Diffusion Sampling.}
We adopt latent diffusion models~\cite{LDM}, where images are encoded into latent representations via an autoencoder $D(E(\cdot))$ and generated through an iterative denoising process~\cite{DDPM,DDIM}. Given a noisy latent $\bm{z}_t$, DDIM~\cite{DDIM} predicts the clean sample as:
\begin{equation}
    \hat{\bm{z}}_{0|t} =
    \frac{1}{\sqrt{\alpha_t}}
    \left(
    \bm{z}_t
    -
    \sqrt{1-\alpha_t}
    \bm{\epsilon}_{\phi}(\bm{z}_t,t)
    \right),
    \label{eq:predicted_z0}
\end{equation}
where $\bm{\epsilon}_{\phi}$ denotes the predicted noise. Guided diffusion sampling further steers the denoising trajectory using task-specific objectives~\cite{UGD,Freedom}. More details are provided in Appendix~\ref{appendix:guided_diffusion}.

\section{Method}

\begin{figure}[!t]
    \centering
    \includegraphics[width=\textwidth]{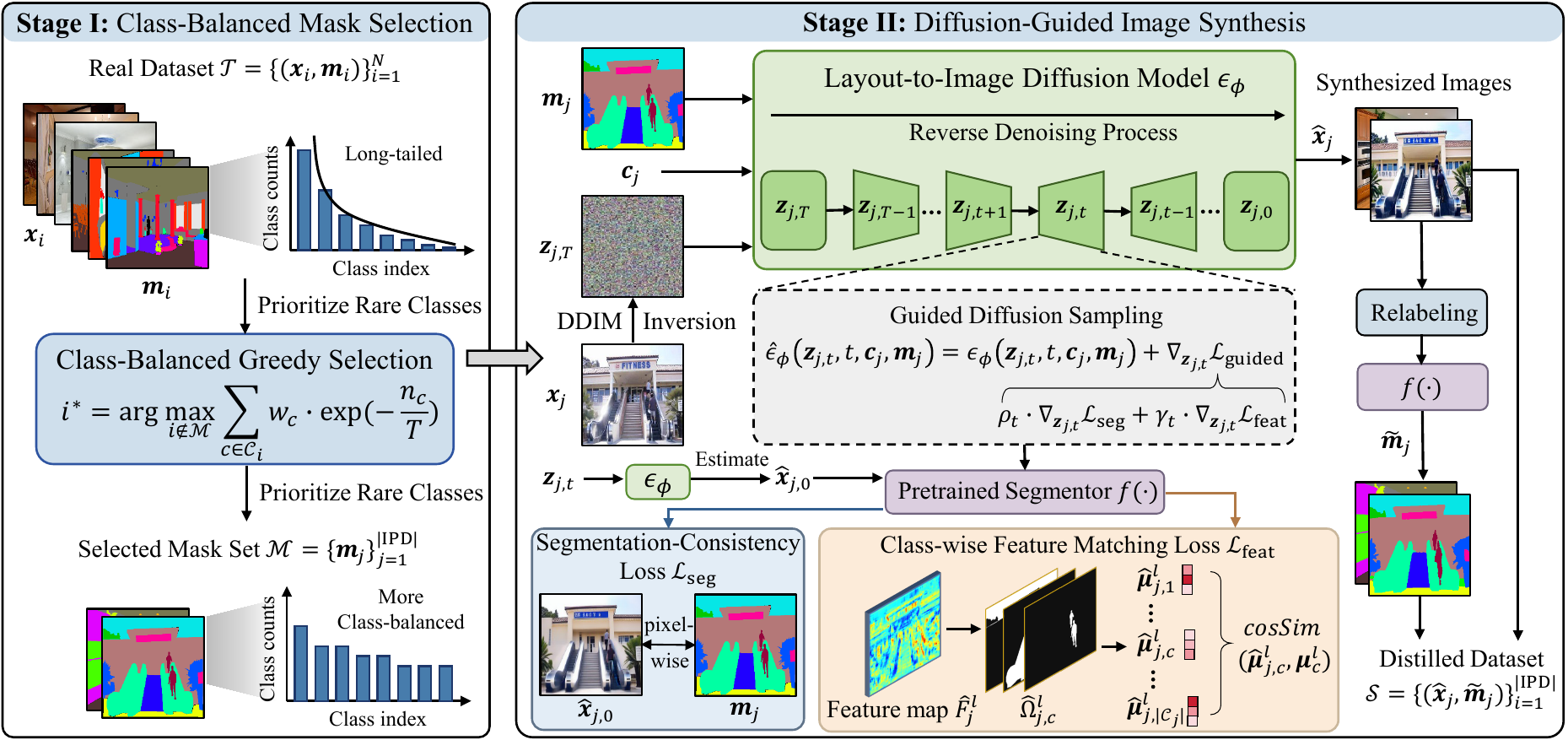}
    \caption{Overview of the \MethodName. In the first stage, we construct a compact mask set that captures long-tailed semantic distributions. In the second stage, we generate layout-aligned images via a layout-to-image diffusion model, further enhanced by guided sampling with segmentation-consistency and class-wise feature matching losses to produce high-quality distilled data.}
    \vspace{-2mm}
    \label{fig:framework}
\end{figure}

To address the challenges of dataset distillation for semantic segmentation, we propose \textbf{\MethodName}, a diffusion-guided framework with a two-stage pipeline, as illustrated in~\Cref{fig:framework}. Specifically, our framework consists of \textit{Class-Balanced Mask Selection} and \textit{Diffusion-Guided Image Synthesis}, which together enable efficient generation of high-quality distilled data.

\subsection{Stage I: Class-Balanced Mask Selection}

\begin{figure}[!t]
    \centering
    \includegraphics[width=\textwidth]{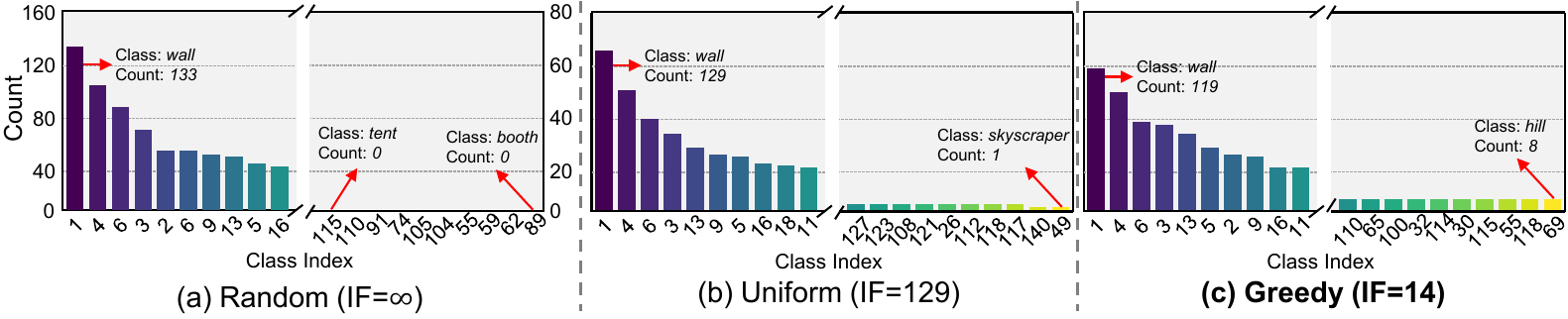}
    \caption{Class distribution comparison under 1\% compression ratio on ADE20K.}
    \vspace{-2mm}
    \label{fig:class distribution comparasion}
\end{figure}

Given the long-tailed class distribution in segmentation datasets (see~\Cref{fig:teaser} (d)), a key challenge is to construct a compact yet representative mask set under a limited distillation budget. A naive strategy that enforces strict class-wise balance (e.g., selecting one sample per class) is suboptimal, as segmentation images typically contain multiple co-occurring classes. Such rigid allocation ignores contextual diversity and inter-class relationships. This issue is illustrated in~\Cref{fig:class distribution comparasion}, which compares different mask selection strategies on ADE20K under a 1\% budget. Random selection severely underrepresents tail classes, with several rare categories never appearing in the selected subset. Rigid uniform selection slightly improves tail coverage by iteratively selecting samples for each class, but still suffers from severe imbalance, yielding an Imbalance Factor (IF) of 129.

To address this, we propose a \textit{Class-Balanced Greedy Selection} strategy that progressively builds a mask subset by jointly considering class rarity and coverage. Let $\mathcal{T}=\{(\bm{x}_i,\bm{m}_i)\}_{i=1}^N$ denote the full dataset, and $\mathcal{C}_i \subset \{0,1,\dots,K-1\}$ be the set of classes present in mask $\bm{m}_i$. Each class $c$ is assigned a weight $w_c$ inversely proportional to its frequency, reflecting its importance under long-tailed distributions. At each step, we select a mask that maximizes the following score:
\begin{equation}
\label{eq:mask_selection}
    i^* = \arg\max_{i \notin \mathcal{M}} \sum_{c \in \mathcal{C}_i} w_c \cdot \exp\left(-\frac{n_c}{T}\right),
\end{equation}
where $\mathcal{M}$ is the current selected set, $n_c$ denotes the coverage count of class $c$, and $T$ is a temperature hyperparameter. The exponential term encourages selecting masks containing underrepresented classes while gradually reducing redundancy as coverage increases.

The selection process is repeated until the distillation budget is reached, yielding the final mask subset $\mathcal{M}=\{\bm{m}_j\}_{j=1}^{|\texttt{IPD}|}$. Compared to rigid per-class sampling, our strategy better exploits the multi-class nature of segmentation images by favoring masks with diverse and complementary semantic information. As shown in~\Cref{fig:class distribution comparasion} (c), our method substantially improves tail coverage, where even the rarest classes appear at least 8 times and the IF is reduced to 14. However, due to inherent class co-occurrence, achieving a perfectly uniform class distribution is infeasible; instead, our method balances diversity and coverage. The full class distribution statistics are provided in Appendix~\ref{appendix:Dataset Distributions}. Detailed complexity and theoretical approximation analysis are provided in Appendix~\ref{appedix:Computational Complexity and Approximation Analysis}.

\subsection{Stage II: Diffusion-Guided Image Synthesis}

\textbf{Layout-to-Image Diffusion Model.}
Given the selected mask set from Stage I, we employ a layout-to-image diffusion model to synthesize images aligned with the semantic layouts. Specifically, we adopt Freestyle~\cite{freestyle}, which enables precise mask-conditioned generation based on a \textit{Stable Diffusion}~\cite{LDM} backbone without introducing additional modules. To reduce domain discrepancy, the model is pretrained on the target dataset $\mathcal{T}$. During sampling, the noise predictor is formulated as $\bm{\epsilon}_{\phi}(\bm{z}_{j,t}, t, \bm{c}_j, \bm{m}_j)$, where $\bm{m}_j$ denotes the input mask and $\bm{c}_j$ is a textual condition constructed by concatenating the class names present in $\bm{m}_j$. This design ensures semantically coherent images that are well aligned with the given layouts.

\textbf{DDIM Inversion.}
Despite effective layout conditioning, we observe that small objects (e.g., \textit{barrel}, \textit{bench}) are occasionally missing in the generated images (see~\Cref{fig:ablation_generation}), likely due to being dominated by large-scale structures during the early denoising. To alleviate this issue, we employ DDIM inversion~\cite{DDIM} to initialize the sampling from a more informative latent. Given a real image corresponding to mask $\bm{m}_j$, we invert it into $\bm{z}_{j,T}$ via the deterministic DDIM trajectory:
\begin{equation}
\bm{z}_{j,t+1} = \sqrt{\alpha_{t+1}} \hat{\bm{z}}_{j,0|t} + \sqrt{1-\alpha_{t+1}} \bm{\epsilon}_{\phi}(\bm{z}_{j,t}, t, \bm{c}_j, \bm{m}_j).
\end{equation}

This initialization provides stronger structural priors, improving the preservation of small objects.

\subsubsection{Guided Diffusion Sampling}

\begin{figure}[!t]
    \centering
    \includegraphics[width=\textwidth]{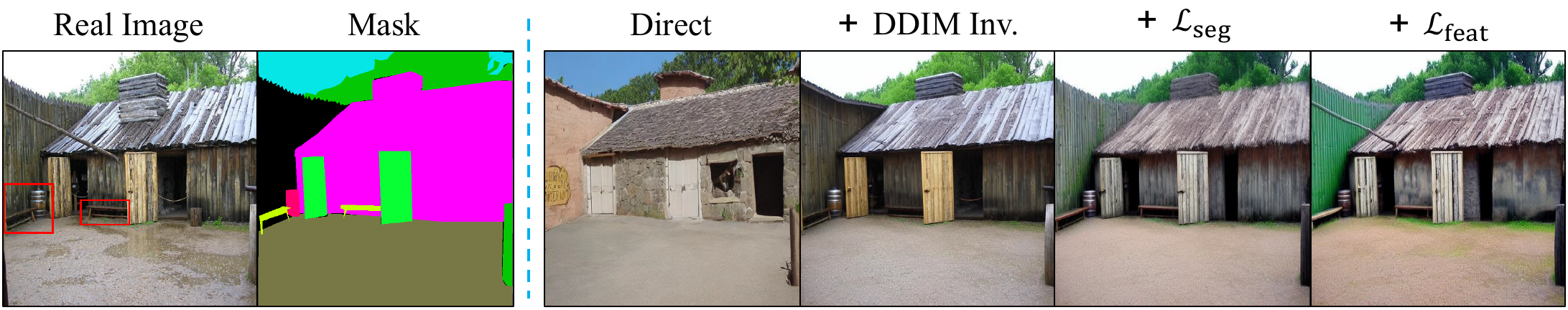}
    \caption{Effect of DDIM Inversion and guided diffusion losses on layout-conditioned image synthesis. We compare generated results under different settings: direct mask-conditioned sampling, with DDIM inversion, and with additional guidance signals using $\mathcal{L}_{\text{seg}}$ and $\mathcal{L}_{\text{feat}}$. Notably, DDIM inversion improves the preservation of small objects (e.g., \textit{barrel} and \textit{bench}), while the proposed guidance further enhances structural completeness and semantic fidelity.}
    \vspace{-4mm}
    \label{fig:ablation_generation}
\end{figure}

While the diffusion model enforces spatial alignment with masks, it does not explicitly guarantee that the synthesized samples are informative for training. To further enhance the training utility of the distilled data, we introduce guided diffusion sampling with two complementary objectives.

\textbf{Segmentation-Consistency Loss.}
We further enforce pixel-wise alignment using a pretrained segmentation model $f(\cdot)$ (readily available from public checkpoints\footnote{\url{https://github.com/open-mmlab/mmsegmentation}}) as guidance. At each step, we estimate the clean latent $\hat{\bm{z}}_{j,0|t}$ via~\Cref{eq:predicted_z0} and decode it into the image space as $\hat{\bm{x}}_{j,0}=D(\hat{\bm{z}}_{j,0|t})$. The $\hat{\bm{x}}_{j,0}$ and $\bm{m}_j$ are then fed into $f$ to compute a standard segmentation loss $\mathcal{L}_{\text{seg}}\big(f(\hat{\bm{x}}_{j,0}), \bm{m}_j\big)$~\cite{mask2former,Segformer}, which encourages consistency between generated images and target masks throughout the denoising process, enhancing class-specific representations (e.g., see the \textit{bench} in~\Cref{fig:ablation_generation}).

\textbf{Class-wise Feature Matching Loss.}
Global BatchNorm statistics~\cite{SRe2L} are insufficient for capturing class-specific features due to spatial averaging. To overcome this, we propose a class-wise feature matching loss that enforces alignment at a finer granularity. We denote by $\bm{F}_i^l \in \mathbb{R}^{H_l \times W_l \times C_l}$ the feature map at the $l$-th stage of $f$ for image $\bm{x}_{i}$. For each class $c$, we resize the mask $\bm{m}_i$ to match the spatial resolution of $\bm{F}_i^l$, and compute the channel-wise mean over its spatial region:
\begin{equation}
\bm{\mu}_{i,c}^l = \frac{1}{|\Omega_{i,c}^l|} \sum_{(u,v)\in \Omega_{i,c}^l} \bm{F}_i^l(u,v),
\end{equation}
where $\Omega_{i,c}^l$ denotes the set of spatial locations belonging to class $c$ and then average over the dataset:
\begin{equation}
\bm{\mu}_c^l = \frac{1}{N_c} \sum_{i: c \in \mathcal{C}_i} \bm{\mu}_{i,c}^l.
\end{equation}

During sampling, we compute $\hat{\bm{\mu}}_{j,c}^l$ from $\hat{\bm{x}}_{j,0}$ in the same manner and define $\mathcal{L}_{\text{feat}}$:
\begin{equation}
\mathcal{L}_{\text{feat}} = \sum_{c \in \mathcal{C}_j} \sum_l \left(1 - cosSim(\hat{\bm{\mu}}_{j,c}^l, \bm{\mu}_c^l) \right),
\end{equation}
which encourages class-specific feature consistency across multiple network stages and $cosSim(\cdot,\cdot)$ denotes the cosine similarity.

\textbf{Design Choice: First-Order Statistics Only.} While SRe2L matches both mean and variance for classification, we align only per-class channel means. This reflects a key difference: in classification, BatchNorm statistics aggregate over single-class images, yielding well-defined per-class distributions; in segmentation, statistics are pooled over mask regions within multi-class scenes where spatial support varies drastically. The dataset-level variance thus conflates genuine intra-class variation with nuisance factors (region size, context). Matching variance would inject artificial heterogeneity that conflicts with $\mathcal{L}_{\text{seg}}$ and the canonical appearance goal. We validate this empirically in \Cref{tab:mean_variance_ablation}.

We incorporate both objectives into the diffusion process by modifying the noise prediction as:
\begin{equation}
\hat{\bm{\epsilon}}_{\phi}(\bm{z}_{j,t},t,\bm{c}_j,\bm{m}_j)
= \bm{\epsilon}_{\phi}(\bm{z}_{j,t},t,\bm{c}_j,\bm{m}_j) \
+ \rho_t \cdot \nabla_{\bm{z}_{j,t}} \mathcal{L}_{\text{seg}} + \gamma_t \cdot \nabla_{\bm{z}_{j,t}} \mathcal{L}_{\text{feat}},
\end{equation}
where $\rho_t$ and $\gamma_t$ control the guidance strength and are normalized to stabilize optimization~\cite{IGD} as:
\begin{equation}
\rho_t = \lambda_{\text{seg}} \cdot \sqrt{1-\alpha_t} \frac{||\bm{\epsilon}_{\phi}(\bm{z}_{j,t},t,\bm{c}_j,\bm{m}_j)||}{||\nabla_{\bm{z}_{j,t}} \mathcal{L}_{\text{seg}}||}, \quad
\gamma_t = \lambda_{\text{feat}} \cdot \sqrt{1-\alpha_t} \frac{||\bm{\epsilon}_{\phi}(\bm{z}_{j,t},t,\bm{c}_j,\bm{m}_j)||}{||\nabla_{\bm{z}_{j,t}} \mathcal{L}_{\text{feat}}||}.
\end{equation}

Overall, $\mathcal{L}_{\text{seg}}$ enforces pixel-level consistency, while $\mathcal{L}_{\text{feat}}$ promotes discriminative, class-aware representations. Notably, both objectives are integrated directly into the sampling process, avoiding costly per-sample pixel optimization. In practice, high-quality distilled samples can be generated within 50 sampling steps ($\sim1$ minute per image), achieving both efficiency and effectiveness.

\textbf{Relabeling.} Following the practice of decoupled methods~\cite{SRe2L}, we further refine the supervision of the synthesized data via relabeling. Given a generated image $\hat{\bm{x}}_j$, we employ the same segmentation model $f(\cdot)$ used during guided sampling to predict its label: 
\begin{equation}
\tilde{\bm{m}}_j = \arg\max f(\hat{\bm{x}}_j).
\end{equation}
The final distilled dataset is thus given by $\mathcal{S}=\{(\hat{\bm{x}}_j,\tilde{\bm{m}}_j)\}_{j=1}^{|\texttt{IPD}|}$. This step corrects potential inconsistencies and provides more informative supervision, leading to improved downstream performance.

\section{Experiments}

\begin{table}[!t]
\centering
\small
\caption{mIoU (\%) on ADE20K~\cite{ade20k} under different distillation ratios. Distilled data are generated using a fixed guidance model (Swin-T for Mask2Former and MiT-B2 for SegFormer).}
\label{tab:ade20k_distill}

\begin{tabular}{c|c|c|c|c|c|c|c}
\toprule

\multirow{2}{*}{Ratio} 
& \multirow{2}{*}{Method}
& \multicolumn{4}{c|}{Mask2Former~\cite{mask2former}} 
& \multicolumn{2}{c}{SegFormer~\cite{Segformer}} \\

\cmidrule(lr){3-6} \cmidrule(lr){7-8}

& & Res-50 & Swin-T & Swin-S & Swin-B & MiT-B2 & MiT-B4 \\

\midrule

\multirow{5}{*}{0.5\%}
& Random   & 9.53{\scriptsize $\pm$0.5} & 9.57{\scriptsize $\pm$0.4} & 10.98{\scriptsize $\pm$0.5} & 13.27{\scriptsize $\pm$0.3} & 12.59{\scriptsize $\pm$0.5} & 15.04{\scriptsize $\pm$0.6} \\
& K-Center & 7.92{\scriptsize $\pm$0.4} & 8.66{\scriptsize $\pm$0.3} & 10.52{\scriptsize $\pm$0.4} & 11.92{\scriptsize $\pm$0.5} & 12.31{\scriptsize $\pm$0.4} & 14.83{\scriptsize $\pm$0.4} \\
& Herding  & 9.75{\scriptsize $\pm$0.5} & 10.50{\scriptsize $\pm$0.4} & 12.47{\scriptsize $\pm$0.4} & 13.63{\scriptsize $\pm$0.3} & 13.30{\scriptsize $\pm$0.4} & 15.78{\scriptsize $\pm$0.3} \\
& Uniform  & \underline{12.54{\scriptsize $\pm$0.2}} & \underline{12.51{\scriptsize $\pm$0.3}} & \underline{15.04{\scriptsize $\pm$0.2}} & \underline{16.15{\scriptsize $\pm$0.4}} & \underline{16.32{\scriptsize $\pm$0.3}} & \underline{19.53{\scriptsize $\pm$0.2}} \\
& \textbf{Ours} & \textbf{15.87{\scriptsize $\pm$0.6}} & \textbf{16.59{\scriptsize $\pm$0.4}} & \textbf{19.45{\scriptsize $\pm$0.5}} & \textbf{22.55{\scriptsize $\pm$0.6}} & \textbf{20.28{\scriptsize $\pm$0.3}} & \textbf{24.45{\scriptsize $\pm$0.5}} \\

\midrule

\multirow{5}{*}{1\%}
& Random   & 12.99{\scriptsize $\pm$0.4} & 13.32{\scriptsize $\pm$0.3} & 15.65{\scriptsize $\pm$0.5} & 18.09{\scriptsize $\pm$0.3} & 16.86{\scriptsize $\pm$0.2} & 19.74{\scriptsize $\pm$0.4} \\
& K-Center & 9.88{\scriptsize $\pm$0.4}  & 11.14{\scriptsize $\pm$0.3} & 13.76{\scriptsize $\pm$0.4} & 15.09{\scriptsize $\pm$0.2} & 14.90{\scriptsize $\pm$0.3} & 18.95{\scriptsize $\pm$0.3} \\
& Herding  & 13.46{\scriptsize $\pm$0.4} & 14.43{\scriptsize $\pm$0.5} & 16.32{\scriptsize $\pm$0.3} & 17.94{\scriptsize $\pm$0.3} & 16.78{\scriptsize $\pm$0.4} & 19.88{\scriptsize $\pm$0.2} \\
& Uniform  & \underline{17.20{\scriptsize $\pm$0.3}} & \underline{17.94{\scriptsize $\pm$0.2}} & \underline{21.42{\scriptsize $\pm$0.3}} & \underline{22.70{\scriptsize $\pm$0.4}} & \underline{20.36{\scriptsize $\pm$0.3}} & \underline{23.48{\scriptsize $\pm$0.4}} \\
& \textbf{Ours} & \textbf{20.96{\scriptsize $\pm$0.3}} & \textbf{21.69{\scriptsize $\pm$0.3}} & \textbf{24.99{\scriptsize $\pm$0.5}} & \textbf{27.48{\scriptsize $\pm$0.5}} & \textbf{23.82{\scriptsize $\pm$0.4}} & \textbf{28.17{\scriptsize $\pm$0.5}} \\

\midrule

\multirow{5}{*}{2\%}
& Random   & 17.50{\scriptsize $\pm$0.3} & 19.16{\scriptsize $\pm$0.4} & 21.70{\scriptsize $\pm$0.4} & 24.00{\scriptsize $\pm$0.3} & 20.72{\scriptsize $\pm$0.3} & 24.53{\scriptsize $\pm$0.2} \\
& K-Center & 13.75{\scriptsize $\pm$0.2} & 14.67{\scriptsize $\pm$0.2} & 17.67{\scriptsize $\pm$0.3} & 20.74{\scriptsize $\pm$0.2} & 19.14{\scriptsize $\pm$0.4} & 22.56{\scriptsize $\pm$0.3} \\
& Herding  & 18.34{\scriptsize $\pm$0.3} & 18.98{\scriptsize $\pm$0.2} & 22.81{\scriptsize $\pm$0.3} & 24.61{\scriptsize $\pm$0.4} & 22.56{\scriptsize $\pm$0.4} & 26.48{\scriptsize $\pm$0.3} \\
& Uniform  & \underline{21.63{\scriptsize $\pm$0.3}} & \underline{23.19{\scriptsize $\pm$0.2}} & \underline{27.59{\scriptsize $\pm$0.3}} & \underline{29.10{\scriptsize $\pm$0.4}} & \underline{25.22{\scriptsize $\pm$0.3}} & \underline{28.46{\scriptsize $\pm$0.4}} \\
& \textbf{Ours} & \textbf{24.67{\scriptsize $\pm$0.4}} & \textbf{26.13{\scriptsize $\pm$0.6}} & \textbf{30.28{\scriptsize $\pm$0.5}} & \textbf{32.63{\scriptsize $\pm$0.7}} & \textbf{28.15{\scriptsize $\pm$0.3}} & \textbf{31.46{\scriptsize $\pm$0.4}} \\

\midrule

\multicolumn{2}{c|}{Full Dataset}
& 47.03{\scriptsize $\pm$0.2} & 48.92{\scriptsize $\pm$0.2} & 51.22{\scriptsize $\pm$0.1} & 52.82{\scriptsize $\pm$0.2} & 45.14{\scriptsize $\pm$0.3} & 48.20{\scriptsize $\pm$0.2} \\

\bottomrule
\end{tabular}
\vspace{-4mm}
\end{table}

\begin{table}[!t]
\centering
\small
\caption{mIoU (\%) on COCO-Stuff~\cite{coco_stuff} under different distillation ratios. Distilled data are generated using a Swin-T guidance model for sampling and relabeling.}
\label{tab:coco_distill}

\begin{tabular}{c|c|c|c|c|c|c}
\toprule
\multirow{2}{*}{Method}
 & \multicolumn{2}{c|}{0.25\%}
 & \multicolumn{2}{c|}{0.5\%}
 & \multicolumn{2}{c}{1\%} \\
\cmidrule(lr){2-3} \cmidrule(lr){4-5} \cmidrule(lr){6-7}
 & Swin-T & Swin-S & Swin-T & Swin-S & Swin-T & Swin-S \\
\midrule
Random   & 15.62{\scriptsize$\pm$0.5} & 18.88{\scriptsize$\pm$0.6} & 21.33{\scriptsize$\pm$0.4} & 24.75{\scriptsize$\pm$0.5} & 26.43{\scriptsize$\pm$0.5} & 29.79{\scriptsize$\pm$0.4} \\
K-Center & 15.11{\scriptsize$\pm$0.4}  & 18.97{\scriptsize$\pm$0.4}  & 20.58{\scriptsize$\pm$0.5}  & 24.58{\scriptsize$\pm$0.4}  & 26.30{\scriptsize$\pm$0.3}  & 30.19{\scriptsize$\pm$0.2}  \\
Herding  & 18.75{\scriptsize$\pm$0.3}  & 22.59{\scriptsize$\pm$0.3}  & 23.50{\scriptsize$\pm$0.2}  & 27.14{\scriptsize$\pm$0.1}  & 28.98{\scriptsize$\pm$0.2}  & 32.50{\scriptsize$\pm$0.3}  \\
Uniform  & \underline{17.54{\scriptsize$\pm$0.4}} & \underline{20.63{\scriptsize$\pm$0.5}} & \underline{22.83{\scriptsize$\pm$0.5}} & \underline{27.54{\scriptsize$\pm$0.3}} & \underline{28.96{\scriptsize$\pm$0.3}} & \underline{32.46{\scriptsize$\pm$0.3}} \\
\textbf{Ours}     & \textbf{22.34{\scriptsize$\pm$0.5}} & \textbf{26.87{\scriptsize$\pm$0.5}} & \textbf{26.16{\scriptsize$\pm$0.6}} & \textbf{30.66{\scriptsize$\pm$0.4}} & \textbf{31.19{\scriptsize$\pm$0.5}}  & \textbf{35.49{\scriptsize$\pm$0.4}}  \\
\midrule
Full Dataset & 46.69{\scriptsize$\pm$0.2} & 48.19{\scriptsize$\pm$0.2} & 46.69{\scriptsize$\pm$0.2} & 48.19{\scriptsize$\pm$0.2} & 46.69{\scriptsize$\pm$0.2} & 48.19{\scriptsize$\pm$0.2} \\
\bottomrule
\end{tabular}
\vspace{-4mm}
\end{table}

\subsection{Experiment Setup}
\label{sec:Experiment Setup}

\textbf{Datasets and Evaluation Models.}
We evaluate \MethodName\ on ADE20K~\cite{ade20k} and COCO-Stuff~\cite{coco_stuff}, with all images resized to $512 \times 512$. ADE20K contains 20,210 training and 2,000 validation images with 150 semantic classes, while COCO-Stuff includes 118,287 training and 5,000 validation images with 182 classes. To evaluate the quality of distilled data, we adopt two representative architectures, Mask2Former~\cite{mask2former} and SegFormer~\cite{Segformer}, with multiple backbones. For each setting, we repeat training five times and report the mean Intersection-over-Union (mIoU) as $\bar{x} \pm \text{std}$.

\textbf{Implementation Details.}
Following common practice in DD, we evaluate compression ratios around 1\%. Specifically, we consider 0.5\%, 1\%, and 2\% on ADE20K, and 0.25\%, 0.5\%, and 1\% on COCO-Stuff. The temperature parameter in mask selection is set to $T = 0.5$. For image synthesis, we use two pretrained checkpoints from FreeStyle~\cite{freestyle} and adopt the DDIM sampler~\cite{DDIM} with 50 denoising steps. The \textit{classifier-free guidance}~\cite{cfg} scale is set to 2.0, while DDIM inversion also use 50 steps with a scale of 1.0. During guided sampling, we set $\lambda_{\text{seg}}=0.05$ and $\lambda_{\text{feat}}=0.2$, and apply guidance at all steps. For segmentation training, we follow the MMSegmentation codebase\footnote{\url{https://github.com/open-mmlab/mmsegmentation}} with minor adjustments to accommodate the limited data size; full details are provided in the Appendix~\ref{appendix:Segmentation Model Training}. All experiments are conducted on a single NVIDIA RTX 6000 Ada GPU (48GB).

\textbf{Baselines.} We adapt representative core-set selection methods from classification~\cite{Dataset_distillation,GM_1}, including Random~\cite{random}, K-center~\cite{K-center}, and Herding~\cite{Herding_1,Herding_2}. Random randomly selects samples from the original training. K-center and Herding are adapted by extracting global image features using a pre-trained Swin-S backbone (from Mask2Former) and measuring distances with the L2 norm~\cite{Fetch_and_forge}. In addition, we include a class-balanced baseline, \textit{Uniform}, which iteratively traverses classes from tail to head and selects one image per class in a round-robin manner until the budget is exhausted. A comparison with more advanced coreset selection methods is provided in Appendix~\ref{appendix:Comparison with Advanced Coreset Selection Methods}.

\subsection{Experimental Results}

\begin{figure}[!t]
    \centering
    \includegraphics[width=\textwidth]{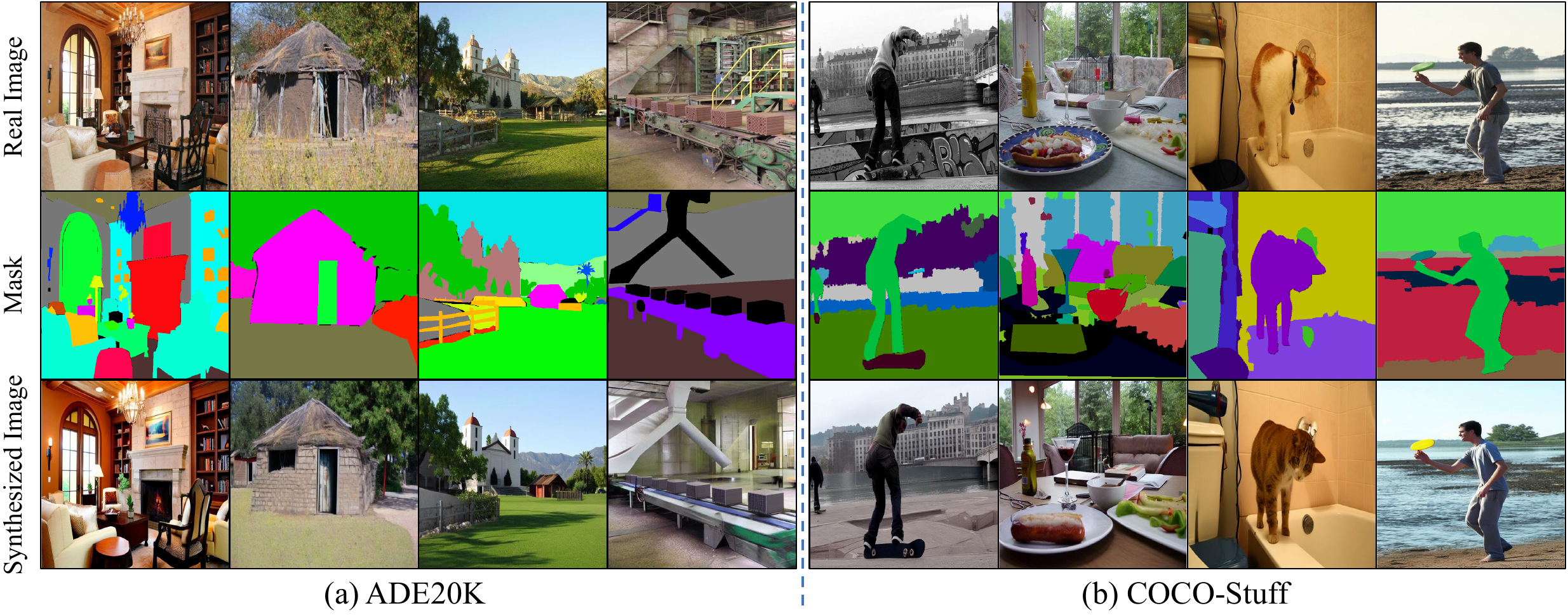}
    \caption{Qualitative visualization of synthesized samples on (a) ADE20K and (b) COCO-Stuff. The synthesized results exhibit strong alignment with the input layouts under both simple and complex scenes, while highlighting class-discriminative features and suppressing irrelevant visual variations.}
    \vspace{-4mm}
    \label{fig:visualization}
\end{figure}

Table~\ref{tab:ade20k_distill} and Table~\ref{tab:coco_distill} report quantitative results on ADE20K and COCO-Stuff. Overall, \MethodName\ consistently outperforms all baselines across architectures and data regimes, demonstrating its effectiveness. On ADE20K, \MethodName\ achieves clear gains at all compression ratios. At 1\%, it surpasses Uniform by 3.57\% on Mask2Former (Swin-S) and 4.69\% on SegFormer (MiT-B4), and remains strong at 0.5\%, improving over Random by up to 9.41\%. Notably, these gains are consistent across backbones despite using a fixed guidance model, indicating strong cross-architecture generalization. In contrast, coreset-based methods perform worse, as they rely on global image-level features and fail to capture dense pixel-level semantics. Although Uniform benefits from class balance, it still lags behind, highlighting the advantage of combining structured mask selection with diffusion-based synthesis. Similar trends appear on COCO-Stuff. At 0.25\%, \MethodName\ outperforms Uniform by 4.80\% and 6.24\% on Swin-T and Swin-S, respectively, with consistent gains at higher ratios (e.g., +3.33\% / +3.12\% at 1\%), showing the robustness of our framework. Despite these improvements, a gap to full-data training remains. For example, on ADE20K, full-data training achieves 51.22\% mIoU on Mask2Former (Swin-S), whereas our best result at 2\% reaches 30.28\%, indicating room for further improvement. Detailed per-class performance breakdowns for rare and difficult categories are provided in Appendix~\ref{appendix:Class-Wise Performance Analysis}.

\textbf{Visualization.} Figure~\ref{fig:visualization} shows qualitative results. The generated images exhibit strong spatial alignment with input masks, with well-preserved object boundaries and spatial structures, demonstrating effective pixel-level consistency. Beyond fidelity, our method produces more \emph{informative} training samples: multiple classes are clearly expressed within a single image, emphasizing discriminative features while suppressing irrelevant variations. Compared to real images with cluttered backgrounds or dataset-specific artifacts, the synthesized samples are cleaner and more canonical, which benefits segmentation training by enhancing class-specific patterns and reducing noise. 
Our approach is particularly beneficial for long-tailed classes. For instance, in ADE20K, the \textit{conveyer belt} category is extremely rare (ratio $<0.005\%$), whereas under a 2\% budget, our method increases its presence to $\sim0.3\%$. The synthesized samples exhibit clear and discriminative \textit{conveyer belt} features with reduced noise, which improves the mIoU of this category from 11.33\% (Uniform) to 30.13\%. Despite these strengths, some limitations remain. Fine-grained structures (e.g., human hands) may exhibit artifacts, likely due to the inherent limitations of the pretrained layout-to-image diffusion model. Improving the underlying generative quality is an important direction for future work.

\subsection{Ablation Study}

\begin{figure}[t]
\centering

\begin{minipage}{0.60\textwidth}
\centering
\small
\setlength{\tabcolsep}{4pt}
\captionof{table}{Ablation study on ADE20K (1\% ratio) using Mask2Former (Swin-T).}

\begin{tabular}{c|c|c|c|c|c|c}
\toprule
Mask Sel. & Diff. & DDIM Inv. & $\mathcal{L}_{\text{seg}}$ & $\mathcal{L}_{\text{feat}}$ & Relabel & mIoU \\
\midrule
$\times$ & $\times$ & $\times$ & $\times$ & $\times$ & $\times$ & 13.32{\scriptsize$\pm$0.8} \\
$\checkmark$ & $\times$ & $\times$ & $\times$ & $\times$ & $\times$ & 18.29{\scriptsize$\pm$0.4} \\
$\checkmark$ & $\checkmark$ & $\times$ & $\times$ & $\times$ & $\times$ & 16.75{\scriptsize$\pm$0.5} \\
$\checkmark$ & $\times$ & $\times$ & $\times$ & $\times$ & $\checkmark$ & 18.34{\scriptsize$\pm$0.5} \\
$\checkmark$ & $\checkmark$ & $\checkmark$ & $\times$ & $\times$ & $\times$ & 17.84{\scriptsize$\pm$0.3} \\
$\checkmark$ & $\checkmark$ & $\checkmark$ & $\checkmark$ & $\times$ & $\times$ & 18.75{\scriptsize$\pm$0.6} \\
$\checkmark$ & $\checkmark$ & $\checkmark$ & $\times$ & $\checkmark$ & $\times$ & 19.36{\scriptsize$\pm$0.5} \\
$\checkmark$ & $\checkmark$ & $\checkmark$ & $\checkmark$ & $\checkmark$ & $\times$ & 20.95{\scriptsize$\pm$0.4} \\
$\checkmark$ & $\checkmark$ & $\checkmark$ & $\checkmark$ & $\checkmark$ & $\checkmark$ & \textbf{21.69{\scriptsize$\pm$0.3}} \\
\bottomrule
\end{tabular}


\label{tab:ablation_components}

\end{minipage}
\hfill
\begin{minipage}{0.37\textwidth}
\centering

\includegraphics[width=\textwidth]{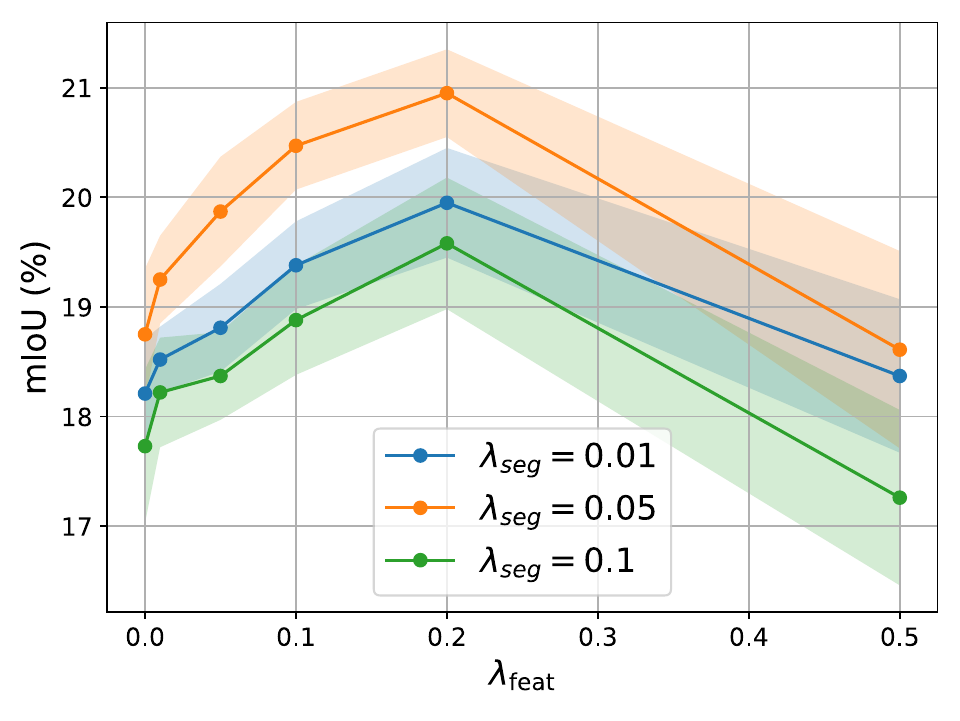}

\caption{Ablation of $\lambda_{\text{seg}}$ and $\lambda_{\text{feat}}$ on ADE20K (1\% ratio) using Mask2Former (Swin-T).}
\label{fig:hyper}

\end{minipage}
\vspace{-4mm}

\end{figure}

\textbf{Effectiveness of Each Component.}
We perform an ablation study on ADE20K at a 1\% ratio, as shown in Table~\ref{tab:ablation_components}, to evaluate the contribution of each component. Starting from a baseline with randomly selected real images (13.32\%), introducing our Stage I mask selection significantly improves performance to 18.29\%, highlighting the importance of constructing a class-balanced subset. Directly applying diffusion-based synthesis without additional guidance degrades performance to 16.75\%, suggesting that naive generation alone is insufficient for producing effective training samples. Incorporating DDIM inversion alleviates this issue (17.84\%) by providing better structural initialization. Introducing the proposed guidance brings consistent gains: $\mathcal{L}_{\text{seg}}$ improves pixel-level alignment (+0.91\%), while $\mathcal{L}_{\text{feat}}$ yields a larger gain (+1.52\%), indicating the importance of class-wise feature consistency. Combining both losses further boosts performance to 20.95\%. Finally, relabeling improves the result to 21.69\%, suggesting that refining supervision helps correct residual inconsistencies in generated data. Overall, each component contributes positively, and their combination achieves the best performance, validating the effectiveness of our design.

\begin{wraptable}{l}{0.50\textwidth}
\centering
\small
\setlength{\tabcolsep}{4pt}
\vspace{-4mm}
\caption{Cross-architecture generalization on ADE20K. Distilled datasets generated with different source architectures (Swin-T or MiT-B2) are evaluated on both student architectures to verify whether synthesis overfits the teacher model.}
\label{tab:cross_arch}
\begin{tabular}{c|c|c|c}
\toprule
Ratio & Source Archi. & Mask2Former & SegFormer \\
\midrule
\multirow{2}{*}{0.5\%}
& Mask2Former & \textbf{16.59{\scriptsize$\pm$0.4}} & 20.27{\scriptsize$\pm$0.4} \\
& SegFormer   & 16.25{\scriptsize$\pm$0.5} & \textbf{20.28{\scriptsize$\pm$0.3}} \\
\midrule
\multirow{2}{*}{1\%}
& Mask2Former & \textbf{21.69{\scriptsize$\pm$0.3}} & 23.57{\scriptsize$\pm$0.3} \\
& SegFormer   & 21.42{\scriptsize$\pm$0.4} & \textbf{23.82{\scriptsize$\pm$0.4}} \\
\bottomrule
\end{tabular}
\end{wraptable}

\textbf{Mean-Only vs. Mean+Variance Matching.} To validate our design choice of using only first-order statistics, we compare mean-only feature matching against a mean+variance variant on ADE20K 1\% across six architectures (Table~\ref{tab:mean_variance_ablation}). Mean-only matching consistently outperforms mean+variance on all six architectures, with improvements ranging from 0.53 to 0.85 mIoU. The consistent superiority indicates that second-order matching is systematically harmful in segmentation, where variance conflates genuine appearance variation with nuisance factors from variable region sizes and contexts.

\begin{table}[!t]
\centering
\small
\caption{Mean-only vs. mean+variance feature matching on ADE20K 1\% (mIoU \%).}
\label{tab:mean_variance_ablation}
\begin{tabular}{l|cccccc}
\toprule
Matching & Res-50 & Swin-T & Swin-S & Swin-B & MiT-B2 & MiT-B4 \\
\midrule
\textbf{Mean only (Ours)} & \textbf{20.96{\scriptsize $\pm$0.3}} & \textbf{21.69{\scriptsize $\pm$0.3}} & \textbf{24.99{\scriptsize $\pm$0.5}} & \textbf{27.48{\scriptsize $\pm$0.5}} & \textbf{23.82{\scriptsize $\pm$0.4}} & \textbf{28.17{\scriptsize $\pm$0.5}} \\
Mean + variance & 20.27{\scriptsize $\pm$0.3} & 21.11{\scriptsize $\pm$0.4} & 24.36{\scriptsize $\pm$0.4} & 26.95{\scriptsize $\pm$0.5} & 23.19{\scriptsize $\pm$0.3} & 27.32{\scriptsize $\pm$0.4} \\
\bottomrule
\end{tabular}
\vspace{-4mm}
\end{table}


\textbf{Cross-architecture Generalization.}
We further evaluate whether the distilled data generalizes across different segmentation architectures. As shown in~\Cref{tab:cross_arch}, datasets distilled with Mask2Former and SegFormer achieve highly consistent performance when evaluated on both models. Under both 0.5\% and 1\% compression ratios, the performance gap between the two source architectures remains small, indicating that synthesis captures intrinsic dataset semantics rather than teacher-specific artifacts. Note that core-set baselines are model-agnostic and thus their cross-architecture performance (reported in \Cref{tab:ade20k_distill}) reflects only student capacity differences rather than generalization. Such strong transferability highlights the robustness of our diffusion-guided distillation framework.

\textbf{Sensitivity of Hyper-Parameters.}
We analyze the effect of $\lambda_{\text{seg}}$ and $\lambda_{\text{feat}}$ on ADE20K at a 1\% ratio (\Cref{fig:hyper}). Overall, introducing moderate feature guidance improves performance, highlighting the importance of class-wise feature alignment. As $\lambda_{\text{feat}}$ increases, mIoU steadily improves until $0.2$, but drops at $\lambda_{\text{feat}}=0.5$, suggesting that overly strong feature constraints may harm the generation quality. For $\lambda_{\text{seg}}$, moderate values perform best. Specifically, $\lambda_{\text{seg}}=0.05$ consistently yields the highest results across different $\lambda_{\text{feat}}$. This indicates that appropriate pixel-level guidance is beneficial, but excessive constraints can limit the flexibility of diffusion sampling. The best performance is achieved at $\lambda_{\text{seg}}=0.05$ and $\lambda_{\text{feat}}=0.2$, which is the default setting. A comprehensive sensitivity analysis of the temperature parameter $T$ in Stage I mask selection is provided in Appendix~\ref{appendix:Sensitivity Analysis of Temperature Parameter}.

\subsection{Efficiency Analysis}







\begin{table}[t]
\centering
\small
\setlength{\tabcolsep}{4pt}
\caption{Efficiency analysis of \MethodName. We report runtime per stage and peak GPU memory.}
\label{tab:efficiency}

\begin{tabular}{c|c|c|c|c|c|c|c|c|c}
\toprule

\multirow{2}{*}{Dataset} &
\multirow{2}{*}{\makecell{Ratio\\(\# Images)}} &
\multirow{2}{*}{\makecell{Mask\\Selection}} &
\multicolumn{2}{c|}{DDIM Inv.} &
\multicolumn{2}{c|}{Synthesis} &
\multicolumn{2}{c|}{Relabel} &
\multirow{2}{*}{Total Time} \\

\cline{4-5}
\cline{6-7}
\cline{8-9}

& & &
Per & Total &
Per & Total &
Per & Total & \\

\midrule

\multirow{3}{*}{ADE20K}
& 0.5\% (101)
& $\sim$1.5 min
& 5 s & $\sim$9 min
& $\sim$50 s & $\sim$1h 20m
& $<1$s & $<1$m
& $\sim$1h 30m \\

& 1\% (202)
& $\sim$3 min
& 5 s & $\sim$18 min
& $\sim$50 s & $\sim$2h 40m
& $<1$s & $<1$m
& $\sim$3h \\

& 2\% (404)
& $\sim$6 min
& 5 s & $\sim$36 min
& $\sim$50 s & $\sim$5h 20m
& $<1$s & $<2$m
& $\sim$6h \\

\midrule

\multirow{3}{*}{COCO-Stuff}
& 0.25\% (295)
& $\sim$8 min
& 5 s & $\sim$26 min
& $\sim$50 s & $\sim$4h
& $<1$s & $<1$m
& $\sim$4h 30m \\

& 0.5\% (591)
& $\sim$16 min
& 5 s & $\sim$52 min
& $\sim$50 s & $\sim$8h
& $<1$s & $<2$m
& $\sim$9h \\

& 1\% (1182)
& $\sim$32 min
& 5 s & $\sim$1h 44m
& $\sim$50 s & $\sim$16h
& $<1$s & $<3$m
& $\sim$18h \\

\midrule

\multicolumn{2}{c|}{Peak Memory}
& ---
& \multicolumn{2}{c|}{$\sim$8GB}
& \multicolumn{2}{c|}{$\sim$28GB}
& \multicolumn{2}{c|}{$\sim$3GB}
& --- \\

\bottomrule
\end{tabular}
\vspace{-4mm}
\end{table}

\Cref{tab:efficiency} summarizes the runtime and memory consumption of each component in \MethodName. Overall, the proposed framework achieves a favorable efficiency–performance trade-off. The \textit{mask selection} stage is highly efficient, requiring only a few minutes even for larger subsets, as it operates on lightweight statistics without involving model training or GPU-intensive computation. The \textit{DDIM inversion} stage introduces negligible overhead, taking only $\sim$5 seconds per sample with moderate memory usage ($\sim$8GB). The primary computational cost lies in the \textit{diffusion-based image synthesis}, which requires $\sim$50 seconds per sample and dominates the total runtime. This is expected, as guided sampling involves gradient backpropagation through the diffusion model, leading to higher memory consumption ($\sim$28GB). Nevertheless, this requirement remains practical on modern GPUs (e.g., 48GB), and can be further reduced via techniques such as gradient checkpointing~\cite{Glad,LD3M}. The \textit{relabeling} stage introduces negligible overhead in both time and memory (less than 1 second per sample and $\sim$3GB), owing to the use of lightweight segmentation models.  Importantly, the overall distillation time scales linearly with the number of samples and remains manageable in practice (e.g., $\sim$3 hours for 1\% ADE20K on a single GPU). Compared to prior dataset distillation methods that rely on costly per-sample pixel optimization or bi-level training, \MethodName\ avoids iterative optimization loops and instead performs one-pass guided generation. As a result, it provides a significantly more efficient and scalable solution for dense prediction tasks.

\section{Conclusion}

This paper presents \MethodName, a diffusion-guided dataset distillation framework for semantic segmentation. It addresses the unique challenges of dense prediction via a two-stage design: (i) a class-balanced greedy selection strategy that constructs a compact and representative mask set for long-tailed distributions, and (ii) layout-to-image diffusion synthesis that generates images aligned with semantic layouts. By integrating segmentation-consistency and class-wise feature matching losses, our method ensures pixel-level alignment and feature fidelity. Extensive experiments on ADE20K and COCO-Stuff demonstrate consistent mIoU improvements over strong baselines across architectures and compression ratios. Overall, this work demonstrates the power of diffusion priors for dataset distillation in semantic segmentation, with substantial room for further improvement.


\begin{ack}


We thank all reviewers for their invaluable feedback, which significantly strengthened this work. This work received no third-party funding or support, and the authors have no competing interests to disclose.

\end{ack}

\bibliographystyle{abbrv}
\bibliography{reference}


\newpage
\appendix

\noindent This appendix provides additional technical details, experimental results, and analyses that complement the main paper. Appendix~\ref{appendix:guided_diffusion} derives the gradient-based guidance mechanism for diffusion-based synthesis and provides the complete algorithm. Appendix~\ref{appendix:Dataset Distributions} visualizes class-distribution statistics for the ADE20K and COCO-Stuff datasets. Appendix~\ref{appendix:Segmentation Model Training} provides extended implementation details. Appendix~\ref{appendix:method_algorithm} presents the complete algorithmic formulation of our two-stage framework. Appendix~\ref{appendix:additional_experiments} presents comprehensive experimental analyses including temperature sensitivity, scaling behavior across compression ratios, comparison with mask-conditioned synthesis methods, fine-grained class-wise performance breakdown for rare and difficult categories, and comparison with advanced coreset selection methods. Appendix~\ref{appendix:Limitations and Future Works} discusses limitations and promising future directions. Appendix~\ref{appendix:Broader Impacts} addresses broader societal impacts. Appendix~\ref{appendix:more_visualizations} provides additional qualitative visualizations on both datasets. Appendix~\ref{appendix:llm_usage} documents large language model usage in manuscript preparation.

\section{More Details on Guided Diffusion Sampling}
\label{appendix:guided_diffusion}

Diffusion models~\cite{DDPM,SDE} learn a parameterized data distribution through a forward noising and reverse denoising process. Given a clean sample $\bm{z}_0$, the forward process gradually perturbs it with Gaussian noise:
\begin{equation}
\bm{z}_t =
\sqrt{\alpha_t}\bm{z}_0
+
\sqrt{1-\alpha_t}\bm{\epsilon},
\quad
\bm{\epsilon}\sim\mathcal{N}(\bm{0},I),
\end{equation}
where $\alpha_t$ controls the noise scale at timestep $t$.

The diffusion model is trained to predict the injected noise
$\bm{\epsilon}_{\phi}(\bm{z}_t,t,\bm{c})$
under condition $\bm{c}$ using a mean squared error objective. During sampling, DDIM~\cite{DDIM} estimates the clean latent:
\begin{equation}
    \hat{\bm{z}}_{0|t} =
    \frac{1}{\sqrt{\alpha_t}}
    \left(
    \bm{z}_t
    -
    \sqrt{1-\alpha_t}
    \bm{\epsilon}_{\phi}(\bm{z}_t,t,\bm{c})
    \right),
\end{equation}
and progressively reconstructs the sample through iterative denoising steps.

In this work, we adopt latent diffusion models~\cite{LDM}, where images are encoded into latent representations $\bm{z}=E(\bm{x})$ and reconstructed using a decoder $D(\cdot)$, significantly improving computational efficiency.

To further steer generation toward desired properties, guided diffusion sampling~\cite{UGD,Freedom} modifies the denoising trajectory using gradients from external objectives. Given a guidance function $f(\cdot)$ and loss function $\ell(\cdot,\cdot)$, the predicted noise is adjusted as:
\begin{equation}
\hat{\bm{\epsilon}}_{\phi}(\bm{z}_t, t, \bm{c})
=
\bm{\epsilon}_{\phi}(\bm{z}_t, t, \bm{c})
+
s(t)\cdot
\nabla_{\bm{z}_t}
\ell\big(f(\bm{z}_t), C\big),
\end{equation}
where $s(t)$ controls the guidance strength and $C$ denotes additional conditions. This formulation enables flexible integration of task-specific objectives into the sampling process.

\section{Dataset Distributions}
\label{appendix:Dataset Distributions}

\begin{figure}[!htbp]
    \centering
    \includegraphics[width=\textwidth]{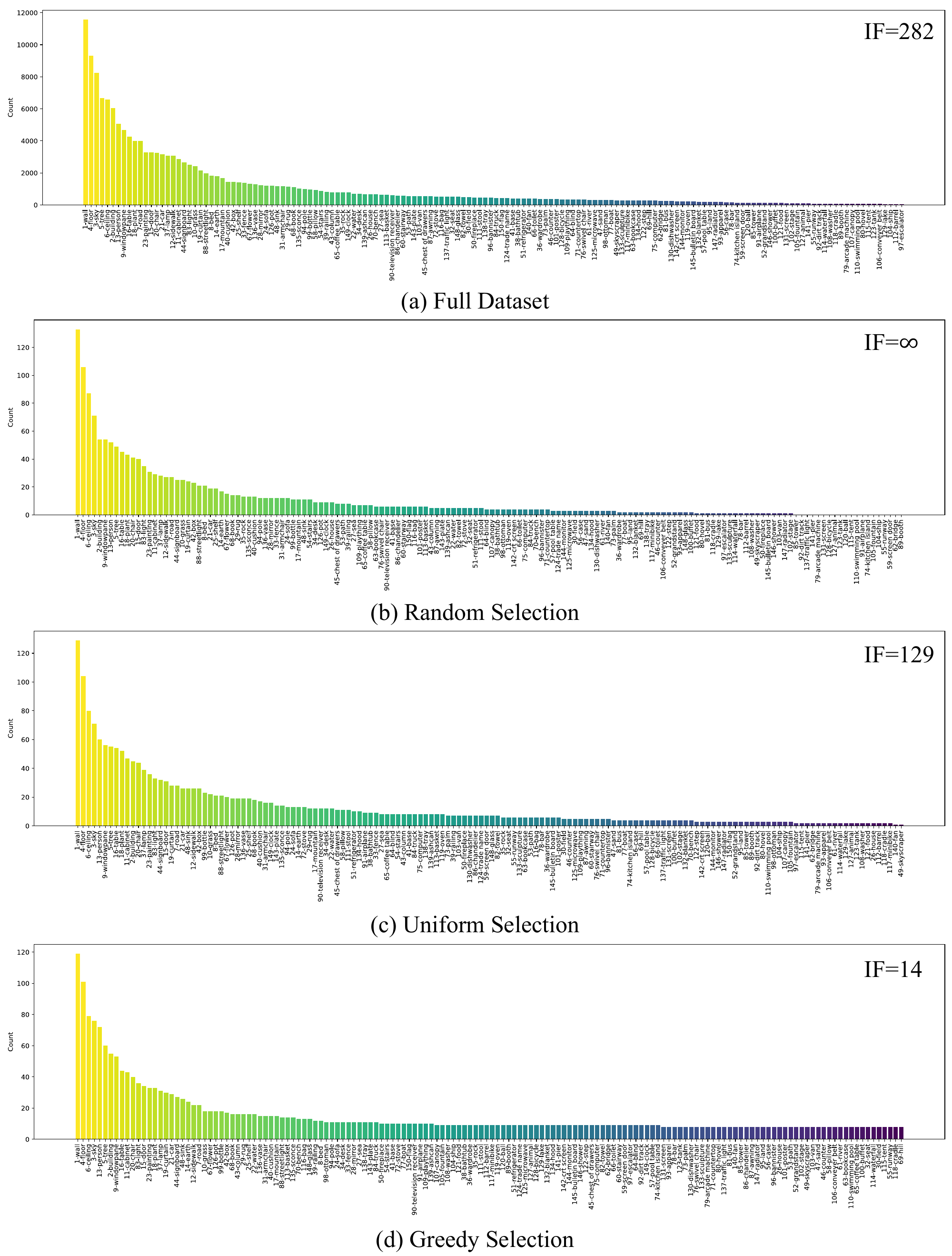}
    \caption{Class distribution comparison on ADE20K under a 1\% selection ratio (202 images). We compare (a) the original dataset, (b) Random selection, (c) Uniform selection, and (d) the proposed Class-Balanced Greedy Selection. The imbalance factor (IF) is reported in each subfigure. Our method significantly reduces class imbalance and improves coverage of tail classes.}
    \label{fig:ade_distributions}
\end{figure}

\begin{figure}[!htbp]
    \centering
    \includegraphics[width=\textwidth]{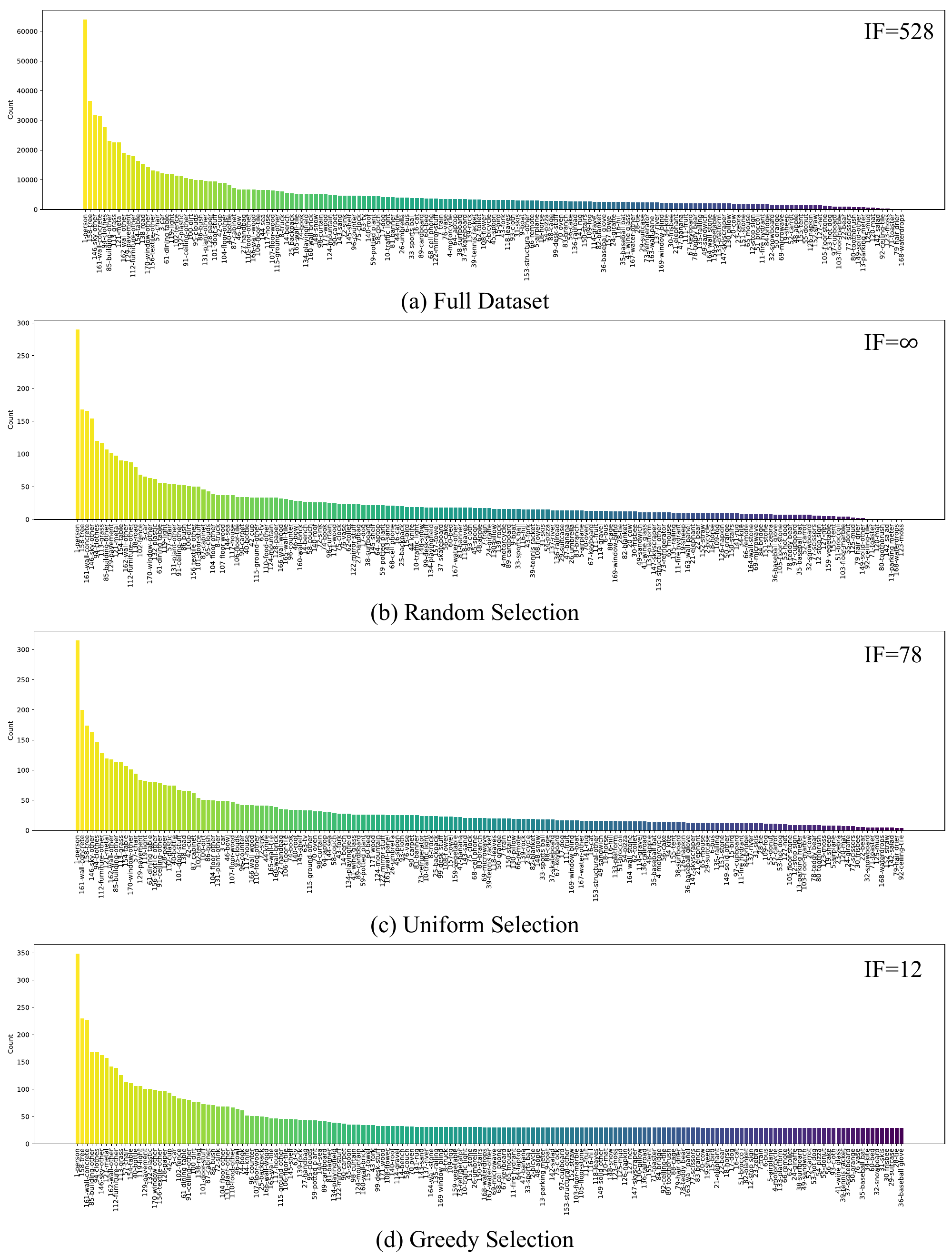}
    \caption{Class distribution comparison on COCO-Stuff under a 0.5\% selection ratio (591 images). We compare (a) the original dataset, (b) Random selection, (c) Uniform selection, and (d) the proposed Class-Balanced Greedy Selection. The imbalance factor (IF) is shown in each subfigure. Our method achieves a much more balanced distribution, especially for rare classes.}
    \label{fig:coco_distributions}
\end{figure}

In this section, we analyze the class distributions of different selection strategies on ADE20K~\cite{ade20k} and COCO-Stuff~\cite{coco_stuff}, as shown in~\Cref{fig:ade_distributions} and~\Cref{fig:coco_distributions}, respectively. For each dataset, we compare four settings: (a) the original dataset, (b) Random selection, (c) Uniform selection, and (d) our Class-Balanced Greedy Selection. The imbalance factor (IF) is reported in each subfigure.

\textbf{ADE20K.}
ADE20K contains 20,210 training images with 150 classes. We fix the compression ratio to 1\%, resulting in 202 selected samples. As shown in~\Cref{fig:ade_distributions} (a), the original dataset exhibits a highly long-tailed distribution with IF=282. Random selection (\Cref{fig:ade_distributions} (b)) fails to cover all classes, leading to an infinite IF. Uniform selection (\Cref{fig:ade_distributions} (c)) partially alleviates the imbalance (IF=129), but tail classes are still severely underrepresented, with only 1--2 samples per class. In contrast, our greedy selection (\Cref{fig:ade_distributions} (d)) significantly improves the balance, reducing IF to 14 and ensuring that even the rarest classes are sampled at least 8 times.

\textbf{COCO-Stuff.}
COCO-Stuff contains 118,287 training images with 182 classes. We use a 0.5\% compression ratio, corresponding to 591 selected samples. The original dataset (\Cref{fig:coco_distributions} (a)) has an even more skewed distribution with IF=528. Similar to ADE20K, Random selection (\Cref{fig:coco_distributions} (b)) leads to missing classes, while Uniform selection (\Cref{fig:coco_distributions} (c)) improves the distribution (IF=78) but still provides limited coverage for tail classes (4--5 samples). Our greedy strategy (\Cref{fig:coco_distributions} (d)) further reduces the imbalance to IF=12, with tail classes appearing at least 29 times, demonstrating substantially improved coverage.

\textbf{Discussion.}
Despite the improved balance, achieving a perfectly uniform class distribution is inherently infeasible in semantic segmentation due to frequent class co-occurrence within images. A single image often contains multiple classes with varying spatial extents, making independent per-class allocation impractical. From a dataset distillation perspective, our strategy is also more principled under extreme compression. Dominant classes (e.g., \textit{person}, \textit{wall}, \textit{sky}) already appear abundantly in the original dataset and contribute less marginal information. Allocating excessive budget to such classes is inefficient. Instead, our method prioritizes underrepresented categories, leading to a more informative and balanced distilled set. This design better utilizes the limited budget and provides more comprehensive supervision, which is crucial for dense prediction tasks like semantic segmentation.

\section{More Implementation Details}
\label{appendix:Segmentation Model Training}

\begin{algorithm}[!t]
\caption{Class-Balanced Greedy Selection}
\label{alg:mask_selection}
\KwIn{Dataset $\mathcal{T}=\{(\bm{x}_i,\bm{m}_i)\}_{i=1}^N$, class sets $\{\mathcal{C}_i\}$, budget $B=|\texttt{IPD}|$, temperature $T$}
\KwOut{Selected mask set $\mathcal{M}$}

Initialize $\mathcal{M} \leftarrow \emptyset$; \quad $n_c \leftarrow 0$ for all classes $c$ \;
Compute class weights $w_c \propto 1/\text{freq}(c)$ \;

\While{$|\mathcal{M}| < B$}{
    \ForEach{$i \notin \mathcal{M}$}{
        Compute score:
        \[
        s_i = \sum_{c \in \mathcal{C}_i} w_c \cdot \exp\left(-\frac{n_c}{T}\right)
        \]
    }
    Select $i^* = \arg\max_i s_i$ \;
    $\mathcal{M} \leftarrow \mathcal{M} \cup \{\bm{m}_{i^*}\}$ \;
    \ForEach{$c \in \mathcal{C}_{i^*}$}{
        $n_c \leftarrow n_c + 1$
    }
}
\Return $\mathcal{M}$
\end{algorithm}

\begin{algorithm}[!t]
\caption{\MethodName: Diffusion-Guided Dataset Distillation for Semantic Segmentation}
\label{alg:main}
\KwIn{Dataset $\mathcal{T}=\{(\bm{x}_i,\bm{m}_i)\}_{i=1}^N$, budget $B$, pretrained layout-to-image diffusion model $\epsilon_{\phi}$, segmentation model $f(\cdot)$}
\KwOut{Distilled dataset $\mathcal{S}$}

\textbf{Stage I: Mask Selection} \;
Obtain mask set $\mathcal{M} \leftarrow \texttt{MaskSelection}(\mathcal{T}, B)$ using Algorithm~\ref{alg:mask_selection} \;

\textbf{Stage II: Diffusion-Guided Image Synthesis} \;
Initialize $\mathcal{S} \leftarrow \emptyset$ \;

\ForEach{$\bm{m}_j \in \mathcal{M}$}{
    Construct condition $\bm{c}_j$ from class names in $\bm{m}_j$ \;

    \textit{// DDIM Inversion} \;
    Obtain latent $\bm{z}_{j,T}$ by inverting corresponding real image $\bm{x}_j$ \;

    \textit{// Guided Sampling} \;
    \For{$t = T$ to $1$}{
        Predict noise $\bm{\epsilon}_{\phi}(\bm{z}_{j,t}, t, \bm{c}_j, \bm{m}_j)$ \;

        Estimate $\hat{\bm{z}}_{j,0|t}$ and decode $\hat{\bm{x}}_{j,0}$ \;

        Compute $\mathcal{L}_{\text{seg}}$ and $\mathcal{L}_{\text{feat}}$ \;

        Update noise with guidance:
        \[
        \hat{\bm{\epsilon}}_{\phi} = \bm{\epsilon}_{\phi} + \rho_t \nabla \mathcal{L}_{\text{seg}} + \gamma_t \nabla \mathcal{L}_{\text{feat}}
        \]

        Sample $\bm{z}_{j,t-1}$ using DDIM update \;
    }

    Decode final image $\hat{\bm{x}}_j = D(\bm{z}_{j,0})$ \;

    \textit{// Relabeling} \;
    Obtain $\tilde{\bm{m}}_j = \arg\max f(\hat{\bm{x}}_j)$ \;

    $\mathcal{S} \leftarrow \mathcal{S} \cup \{(\hat{\bm{x}}_j, \tilde{\bm{m}}_j)\}$ \;
}

\Return $\mathcal{S}$
\end{algorithm}

For downstream evaluation, we train segmentation models using the official MMSegmentation codebase.\footnote{\url{https://github.com/open-mmlab/mmsegmentation}} Specifically, we adopt Mask2Former~\cite{mask2former} and SegFormer~\cite{Segformer} as the evaluation architectures. For training on the full dataset, we follow the official configurations provided by MMSegmentation, such as \texttt{mask2former\_r50\_8xb2-160k\_ade20k-512x512.py} and \texttt{segformer\_mit-b2\_8xb2-160k\_ade20k-512x512.py}. All models are retrained on a server with 8 NVIDIA GeForce RTX 3090 GPUs to ensure a fair comparison. For training on the distilled datasets, we adjust the training schedule to account for the significantly reduced data size. Specifically, we reduce the number of training iterations from 160k to 20k while keeping all other hyperparameters unchanged. This modification ensures efficient training while maintaining consistency with the original setup. Code is available at \url{https://github.com/zwj084/D3S2}.


\section{Method Algorithm}
\label{appendix:method_algorithm}

\subsection{Stage I: Class-Balanced Greedy Selection}

Algorithm~\ref{alg:mask_selection} presents our proposed Class-Balanced Greedy Selection. It progressively constructs a representative mask subset by balancing class rarity and coverage. The class weights prioritize underrepresented categories, while the exponential decay term discourages repeatedly selecting already well-covered classes. This dynamic trade-off enables the selection of diverse and complementary samples under a limited budget. Unlike rigid class-wise sampling, the proposed strategy naturally leverages multi-class co-occurrence in segmentation data, resulting in a more balanced yet contextually rich mask set.

\subsubsection{Computational Complexity and Approximation Analysis}
\label{appedix:Computational Complexity and Approximation Analysis}

While the greedy selection in Algorithm~\ref{alg:mask_selection} is conceptually straightforward, its theoretical justification is rigorous. We show that the algorithm greedily maximizes a monotone submodular objective function, yielding provable approximation guarantees.

\textbf{Submodular Formulation.} Define $\gamma := e^{-1/T}$ and let $n_c(\mathcal{M})$ denote the coverage count of class $c$ in set $\mathcal{M}$. We introduce the \textit{soft class-weighted coverage} objective:
\begin{equation}
F(\mathcal{M}) = \sum_c w_c \cdot \phi(n_c(\mathcal{M})), \quad \text{where} \quad \phi(n) = \frac{1-\gamma^n}{1-\gamma},
\end{equation}
This formulation assigns diminishing returns to successive appearances of class $c$: the first occurrence contributes $w_c$, the second $w_c\gamma$, and so on. The marginal gain of adding a mask $i$ to $\mathcal{M}$ is precisely:
\begin{equation}
\Delta_i F(\mathcal{M}) = F(\mathcal{M} \cup \{i\}) - F(\mathcal{M}) = \sum_{c \in \mathcal{C}_i} w_c \gamma^{n_c(\mathcal{M})} = s_i,
\end{equation}
which matches~\Cref{eq:mask_selection} in the main paper. Thus, Algorithm~\ref{alg:mask_selection} is standard greedy maximization of $\max_{|\mathcal{M}| \le B} F(\mathcal{M})$.

\textbf{Submodularity.} The objective $F$ is monotone and submodular for every $T > 0$. Monotonicity follows immediately since $\phi(n)$ is non-decreasing in $n$. Submodularity holds because $\gamma^n$ is a non-increasing sequence in $n$, which ensures that marginal gains $\Delta_i F(\mathcal{M})$ can only decrease as $\mathcal{M}$ grows. Formally, for any $\mathcal{M} \subseteq \mathcal{M}'$ and item $i \notin \mathcal{M}'$:
\begin{equation}
\Delta_i F(\mathcal{M}') = \sum_{c \in \mathcal{C}_i} w_c \gamma^{n_c(\mathcal{M}')} \le \sum_{c \in \mathcal{C}_i} w_c \gamma^{n_c(\mathcal{M})} = \Delta_i F(\mathcal{M}).
\end{equation}

\textbf{Approximation Guarantee.} Two important consequences follow: \textbf{(1) NP-hardness.} For $w_c \equiv 1$ (uniform class weights) and small $T$ (negligible decay), the problem $\max_{|\mathcal{M}| \le B} F(\mathcal{M})$ reduces to the Maximum Coverage problem, which is known to be NP-hard~\cite{Karp72,Feige98}. \textbf{(2) Optimal Polynomial-Time Approximation.} By the classical result of Nemhauser–Wolsey–Fisher~\cite{NWF1978}, the greedy algorithm achieves $(1-1/e) \approx 0.632$ of the optimal value for maximizing any monotone submodular function subject to a cardinality constraint. Moreover, by Feige~\cite{Feige98}, this $(1-1/e)$ ratio is the best guarantee any polynomial-time algorithm can achieve unless P$=$NP. Therefore, our Algorithm~\ref{alg:mask_selection} is not only simple but also theoretically optimal in the context of polynomial-time approximations.

\textbf{Computational Complexity.} The cost of Algorithm~\ref{alg:mask_selection} is $O(B \cdot N \bar{K})$, where $B$ is the distillation budget, $N$ is the dataset size, and $\bar{K}$ is the average number of classes per image. Since mask selection operates only on lightweight statistics (class frequencies and coverage counts) without model training, it completes in a few CPU-minutes even for large datasets.

\subsection{Stage II: Diffusion-Guided Image Synthesis}

Algorithm~\ref{alg:main} summarizes the overall pipeline of \MethodName. The framework first constructs a representative mask set via class-balanced selection (Stage I), and then synthesizes high-quality training samples through diffusion-based generation with guidance (Stage II). Compared to existing dataset distillation approaches, \MethodName\ offers several advantages. First, it explicitly accounts for long-tailed distributions through adaptive mask selection with theoretical guarantees. Second, it avoids costly pixel-level optimization by integrating task-specific objectives directly into the diffusion sampling process. Finally, the combination of segmentation-consistency and class-wise feature matching enables the generation of structurally aligned and semantically discriminative samples, leading to strong performance and generalization across architectures.

\textbf{Synthesis vs. Refinement.} While we term Stage II \textit{synthesis}, it is more precisely understood as \textit{refinement}. Each selected mask $\bm{m}_j$ comes paired with a corresponding real image $\bm{x}_j$ from the original dataset. Via DDIM inversion, we project this real image into latent space as the starting point for denoising, rather than sampling from random noise. Without our guidance losses, the deterministic DDIM trajectory would return close to the original image. Our two objectives—$\mathcal{L}_{\text{seg}}$ for pixel-level alignment and $\mathcal{L}_{\text{feat}}$ for class-wise feature consistency—then guide refinement away from this natural point. The endpoint is thus neither the original image nor an unrestricted generation, but a conditioned refinement that balances fidelity to the mask with dataset-level feature statistics. This perspective clarifies how Stage II operates: we preserve the real image as an informative prior and refine it toward improved training utility.

\section{Additional Experimental Results}
\label{appendix:additional_experiments}

\subsection{Sensitivity Analysis of Temperature Parameter $T$}
\label{appendix:Sensitivity Analysis of Temperature Parameter}

\textbf{Motivation.} The class-balanced greedy selection (Algorithm~\ref{alg:mask_selection}) relies on a temperature parameter $T$ that controls the trade-off between prioritizing underrepresented classes and maintaining sample diversity. To justify our choice of $T=0.5$ and demonstrate robustness, we provide a comprehensive sensitivity analysis across a $50\times$ range of values. Crucially, $T=0.5$ was selected by minimizing the Imbalance Factor (IF) of the selected mask subset—a criterion that uses only Stage-I mask statistics, requiring no synthesis, student training, or test labels. This ensures $T$ was never tuned on downstream performance and can be reproduced on new datasets at negligible cost.

\textbf{Experimental Setup.} We evaluate six temperature values $T \in \{0.1, 0.2, 0.5, 1.0, 2.0, 5.0\}$ on ADE20K at 1\% compression (202 images) using Mask2Former (Swin-T). For each $T$, we report: (1) Imbalance Factor (IF) and minimum per-class count from Stage I; (2) Stage-I-only mIoU—training on the selected \textit{real} image-mask pairs, isolating selection quality from synthesis; and (3) full \MethodName\ mIoU with diffusion-guided synthesis. All results are averaged over 5 runs.

\begin{table}[!t]
\centering
\caption{Sensitivity analysis of temperature parameter $T$ on ADE20K at 1\% compression (202 images) using Mask2Former (Swin-T). IF: Imbalance Factor (lower indicates better balance). Min: minimum per-class count (higher indicates better coverage). Uniform baseline achieves 17.94 mIoU.}
\label{tab:temperature_sensitivity}
\begin{tabular}{c|cc|c|c}
\toprule
$T$ & IF$\downarrow$ & Min$\uparrow$ & Stage-I-only & Full \MethodName\ \\
\midrule
0.1 & 22 & 6 & 18.04{\scriptsize $\pm$0.1} & 21.14{\scriptsize $\pm$0.4} \\
0.2 & 17 & 7 & 18.15{\scriptsize $\pm$0.1} & 21.38{\scriptsize $\pm$0.3} \\
\textbf{0.5} & \textbf{14} & \textbf{8} & \textbf{18.29{\scriptsize $\pm$0.1}} & \textbf{21.69{\scriptsize $\pm$0.3}} \\
1.0 & 18 & 7 & 18.19{\scriptsize $\pm$0.1} & 21.45{\scriptsize $\pm$0.2} \\
2.0 & 18 & 7 & 18.17{\scriptsize $\pm$0.1} & 21.40{\scriptsize $\pm$0.3} \\
5.0 & 25 & 5 & 18.05{\scriptsize $\pm$0.2} & 21.15{\scriptsize $\pm$0.4} \\
\bottomrule
\end{tabular}
\end{table}

\textbf{Results and Analysis.} Four key patterns emerge from~\Cref{tab:temperature_sensitivity}: \textbf{(1) Robustness across wide range.} Across the $50\times$ range, full-pipeline mIoU spans only 0.55 points (21.14–21.69) with consistently low variance ($\sigma \leq 0.4$), while Stage-I-only mIoU spans merely 0.25 points (18.04–18.29) with minimal variance ($\sigma \leq 0.2$). Within the reasonable range $T \in [0.2, 2.0]$, the spread narrows to 0.31 mIoU, demonstrating that performance is insensitive to the specific choice of $T$. \textbf{(2) Consistent superiority over baselines.} Every tested $T$ outperforms the strongest baseline (Uniform, 17.94 mIoU) by substantial margins. Even the worst configurations ($T=0.1$ and $T=5.0$) exceed Uniform by +3.20 and +3.21 mIoU, respectively. Stage-I-only results surpass Uniform for \textit{all} $T$ values ($\geq$18.04), confirming that our method's advantage is robust and does not depend on careful tuning. \textbf{(3) Optimality of $T=0.5$.} The default choice $T=0.5$ simultaneously achieves the lowest IF (14), highest minimum per-class count (8), and best mIoU for both Stage-I-only (18.29$\pm$0.1) and full pipeline (21.69$\pm$0.3). This choice also exhibits the smallest or near-smallest variance, providing both point-estimate optimality and stability benefits. \textbf{(4) Behavior at extremes.} Performance degrades predictably at both extremes due to the exponential decay term $\exp(-n_c/T)$: as $T \to 0$, it approaches a binary indicator (never-covered vs. covered); as $T \to \infty$, it approaches 1, reducing the score to a static rarity sum. Both extremes diminish the dynamic balance between class rarity and coverage, yielding suboptimal and less stable selection.

\textbf{Discussion.} The robustness across $T$ stems from our selection mechanism: class weights $w_c \propto 1/\text{freq}(c)$ span over two orders of magnitude (original IF=282), dominating the selection score $s_i = \sum_{c \in \mathcal{C}_i} w_c \exp(-n_c/T)$. $T$ modulates only how quickly covered classes are discounted, providing fine-grained control without drastically altering the core prioritization of rare classes. Our IF-based selection criterion naturally identifies the optimal $T$ without requiring downstream task evaluation, making it reproducible and transferable to new datasets at negligible cost.

\subsection{Scaling Experiment: Budget vs. Full-Data Performance}

\textbf{Motivation.} To quantify the regime where \MethodName\ provides maximum value, we evaluate scaling behavior across compression ratios from 0.5\% to 20\% of ADE20K (101 to 4,042 training images out of 20,210 total). This experiment characterizes how our advantage over baselines changes as the data budget increases, validating our design focus on extreme compression ratios.

\textbf{Experimental Setup.} We compare \MethodName\ against Random selection and Uniform across seven budget ratios: 0.5\%, 1\%, 2\%, 5\%, 10\%, 15\%, and 20\%. All experiments use a consistent configuration: $T=0.5$, Swin-T for guidance and relabeling, and identical downstream training protocols with Mask2Former (Swin-T). Results are reported as means over 5 runs, with the 0.5\%, 1\%, and 2\% points corresponding to the experiments from~\Cref{tab:ade20k_distill} in the main paper.

\begin{figure}[!h]
\centering
\includegraphics[width=\textwidth]{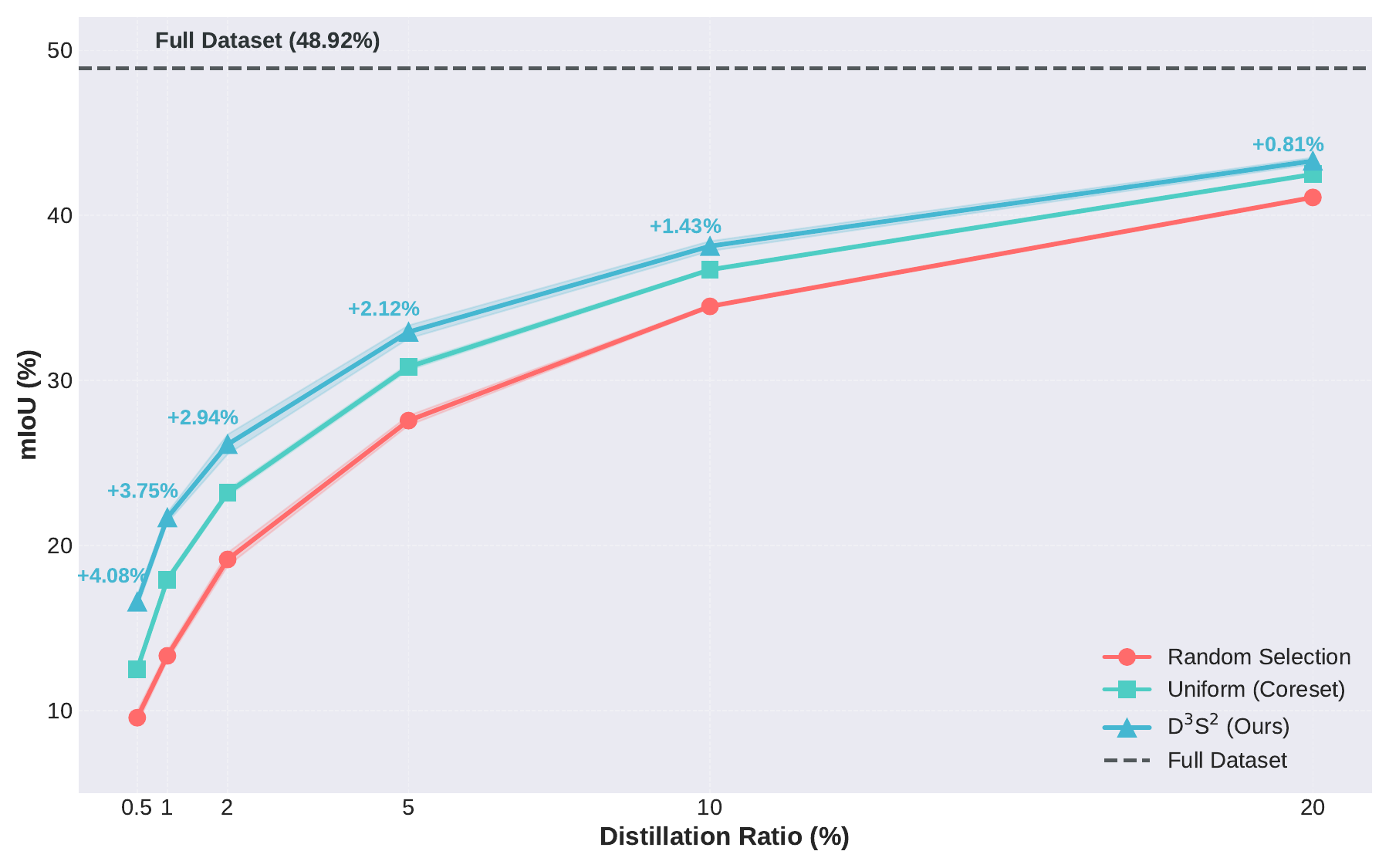}
\caption{Scaling behavior on ADE20K (Mask2Former Swin-T) across compression ratios from 0.5\% to 20\%. \MethodName\ consistently outperforms Random and Uniform across all budgets, with advantages decreasing monotonically from +4.08\% at 0.5\% to +0.81\% at 20\%.}
\label{fig:scaling_experiment}
\end{figure}

\textbf{Results and Analysis.} Three key findings emerge from~\Cref{fig:scaling_experiment}: \textbf{(1) Maximum advantage in low-budget regime.} The margin over Uniform selection decreases monotonically: +4.08\% (0.5\%) $\to$ +3.75\% (1\%) $\to$ +2.94\% (2\%) $\to$ +2.12\% (5\%) $\to$ +1.43\% (10\%) $\to$ +0.81\% (20\%). This demonstrates that \MethodName\ delivers the largest gains precisely where data constraints are most severe—the regime that motivates dataset distillation. \textbf{(2) Bounded gap to full data.} At 20\% compression, our method achieves 43.30\% mIoU compared to full-data performance of 48.92\%, leaving a gap of 5.62 percentage points. This plateau is theoretically expected: dataset distillation compresses a large-scale distribution into a compact surrogate, and once real data coverage becomes sufficient (at higher ratios), any distillation method must asymptotically approach and eventually fall behind real images. This phenomenon has been observed in other dense-prediction condensation work~\cite{Fetch_and_forge}. \textbf{(3) Design positioning validation.} Our focus on 0.25\%–2\% compression ratios (where 101 images at 0.5\% roughly match Random's performance with 404 images) aligns with the distillation objective of achieving 4× data savings in the regime where our advantage is greatest. The scaling curves confirm this is not an arbitrary choice but reflects the natural operating point where distilled data provides maximum value compared to alternatives.

\subsection{Comparison with Mask-Conditioned Synthesis Methods}

\textbf{Motivation.} While \MethodName\ shares its synthesis mechanism with mask-conditioned image generation methods, the objectives differ fundamentally. Such methods typically perform \textbf{Generative Data Augmentation (GDA)}, enriching datasets at modest expansion ratios (1:5 to 1:20) to train \textit{together with} the full real dataset. In contrast, our setting is \textbf{Generative Dataset Distillation (GDD)}, operating at extreme compression ratios (100:1 to 1000:1) where synthesized samples \textit{replace} rather than augment real data. This inversion fundamentally changes design requirements: expansion affords redundancy where weak samples are harmless since real data remains available; distillation cannot, and must make every sample carry dataset-level information.

\textbf{Compared Method.} We compare against FreeMask~\cite{freemask}, a representative mask-conditioned synthesis method for generating segmentation training data. FreeMask employs a three-stage pipeline: (1) selecting diverse layouts from real data and generating synthetic images conditioned on these layouts using pretrained diffusion models; (2) filtering noisy synthetic regions by comparing generated images with their conditioning masks; and (3) re-sampling synthetic images based on mask-level hardness to improve data quality. For fair comparison, we use identical generator checkpoints (FreeStyleNet~\cite{freestyle} fine-tuned on ADE20K) and provide FreeMask with \textit{exactly our Stage-I selected masks}, so the pipelines differ only in Stage II synthesis objectives. We implement stages (1) and (2) of FreeMask—layout-conditioned generation and noise filtering—but exclude the hardness-based re-sampling (stage 3) as it requires additional forward passes and iterative refinement orthogonal to the core synthesis mechanism being compared.

\textbf{Experimental Setup.} We evaluate on ADE20K at a 1\% compression ratio (202 images) across six backbone architectures spanning both Mask2Former (Res-50, Swin-T, Swin-S, Swin-B) and SegFormer (MiT-B2, MiT-B4). All methods use identical Stage-I masks selected by our method, Swin-T for guidance during synthesis, and the same downstream training protocols. We report FreeMask results both without and with relabeling to isolate the contribution of mask-image alignment. Uniform selection serves as the real-image baseline. All results are reported as means±std over 5 runs, with ours corresponding to the experiments from~\Cref{tab:ade20k_distill}.

\begin{table}[!t]
\centering
\caption{Comparison with FreeMask on ADE20K at 1\% compression (202 images, mIoU \%). All methods use identical Stage-I selected masks, differing only in Stage II synthesis objectives.}
\label{tab:freemask_comparison}
\begin{tabular}{l|c|cccccc}
\toprule
Method & Relabel & Res-50 & Swin-T & Swin-S & Swin-B & MiT-B2 & MiT-B4 \\
\midrule
Uniform & -- & 17.20{\scriptsize $\pm$0.3} & 17.94{\scriptsize $\pm$0.2} & 21.42{\scriptsize $\pm$0.3} & 22.70{\scriptsize $\pm$0.4} & 20.36{\scriptsize $\pm$0.3} & 23.48{\scriptsize $\pm$0.4} \\
\midrule
FreeMask & $\times$ & 15.09{\scriptsize $\pm$0.1} & 15.42{\scriptsize $\pm$0.4} & 18.29{\scriptsize $\pm$0.3} & 21.95{\scriptsize $\pm$0.2} & 19.60{\scriptsize $\pm$0.2} & 21.77{\scriptsize $\pm$0.2} \\
FreeMask & $\checkmark$ & 17.27{\scriptsize $\pm$0.2} & 18.10{\scriptsize $\pm$0.3} & 21.62{\scriptsize $\pm$0.2} & 24.76{\scriptsize $\pm$0.2} & 21.29{\scriptsize $\pm$0.2} & 24.90{\scriptsize $\pm$0.3} \\
\midrule
\textbf{Ours} & $\checkmark$ & \textbf{20.96{\scriptsize $\pm$0.3}} & \textbf{21.69{\scriptsize $\pm$0.3}} & \textbf{24.99{\scriptsize $\pm$0.5}} & \textbf{27.48{\scriptsize $\pm$0.5}} & \textbf{23.82{\scriptsize $\pm$0.4}} & \textbf{28.17{\scriptsize $\pm$0.5}} \\
\bottomrule
\end{tabular}
\end{table}

\textbf{Results and Analysis.} Three patterns emerge from~\Cref{tab:freemask_comparison}: \textbf{(1) GDA fails at extreme compression.} Without relabeling, FreeMask falls \textit{below} the Uniform baseline across all six backbones by --0.75to --3.13 mIoU, meaning its synthetic images train worse than the real images whose masks produced them. With relabeling, FreeMask barely edges past Uniform by only +0.16 to +2.06 mIoU (mean +0.81), demonstrating that a GDA pipeline at 100:1 compression barely justifies generation at all. In contrast, our method surpasses Uniform by +3.46 to +4.78 mIoU (mean +4.00). \textbf{(2) Alignment as the primary bottleneck.} Relabeling lifts FreeMask by up to +3.33 mIoU but our pipeline by only +0.74 mIoU (from~\Cref{tab:ablation_components}), indicating FreeMask's generated images drift significantly from the masks used as ground truth. Our $\mathcal{L}_{\text{seg}}$ enforces alignment \textit{during} sampling rather than via post-hoc correction, explaining the smaller relabeling gain. \textbf{(3) Dataset-level statistics matter.} The remaining 2.5--4.2 mIoU advantage after accounting for alignment is attributable to $\mathcal{L}_{\text{feat}}$, which matches dataset-level feature statistics aggregated over real training images. A GDA objective has no reason to match such statistics, as expansion never requires one sample to be representative—redundancy across the augmented set and retention of real data make individual sample quality less critical. In distillation, every sample must encode dataset-level information to compensate for 100:1 compression, necessitating guidance objectives beyond per-image mask conditioning.

\begin{figure}[!h]
\centering
\includegraphics[width=\textwidth]{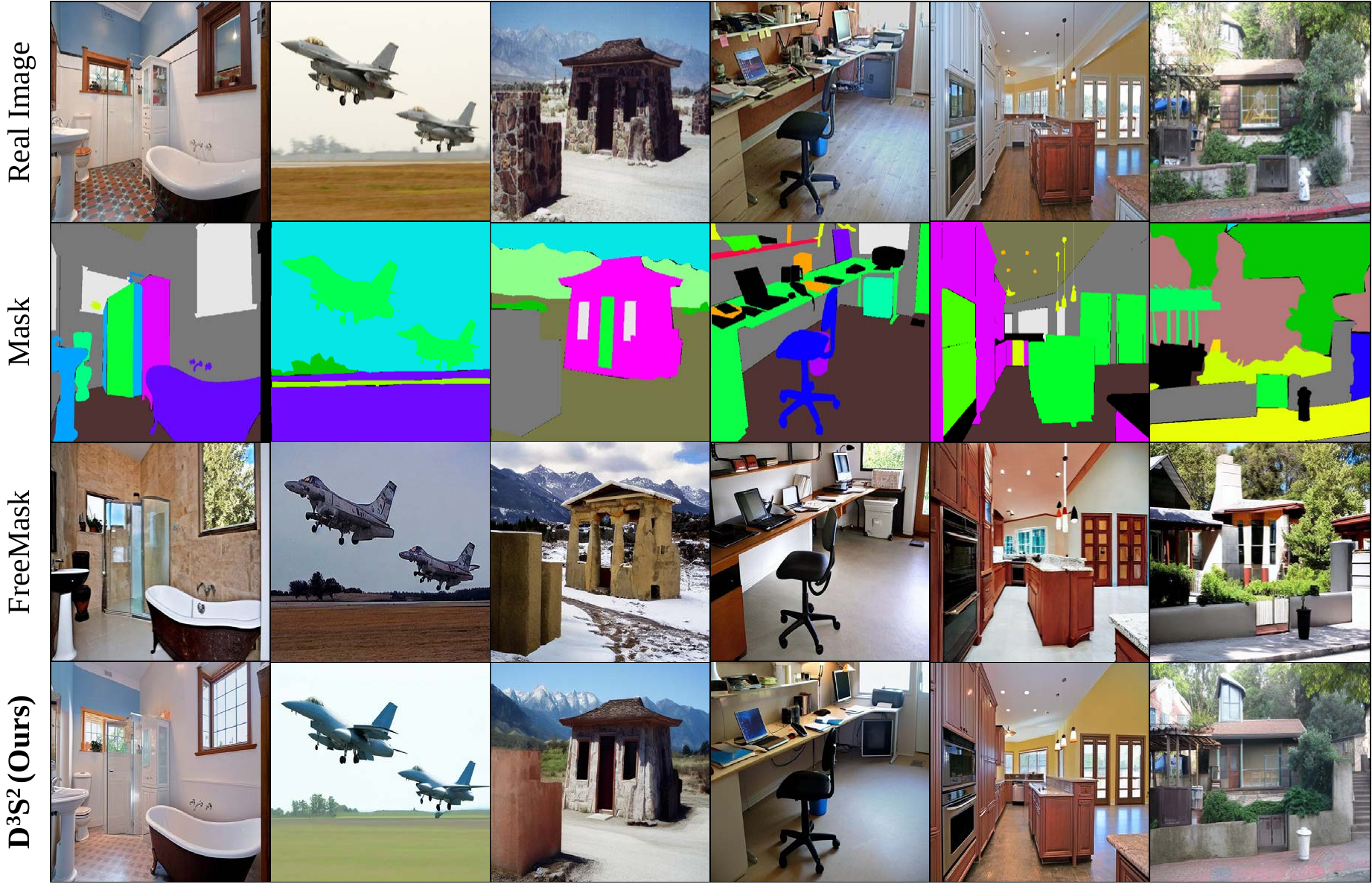}
\caption{Qualitative comparison between FreeMask and \MethodName\ on ADE20K. For each example, we show: real image, segmentation mask, FreeMask synthesis, and our synthesis. Our method demonstrates superior mask-image alignment, particularly for small objects, and generates textures that better capture class-intrinsic attributes.}
\label{fig:compare_freemask}
\end{figure}

\textbf{Qualitative Analysis.} ~\Cref{fig:compare_freemask} visually corroborates the quantitative findings: \textbf{(1) Stronger spatial alignment.} Our synthesized images adhere more faithfully to input masks compared to FreeMask, especially for small and thin objects where precise boundaries are critical. FreeMask's generations exhibit noticeable spatial drift—objects appear displaced, truncated, or with incorrect boundaries—explaining why relabeling provides large gains (+3.33 mIoU) for FreeMask but minimal gains (+0.74 mIoU) for ours. Our $\mathcal{L}_{\text{seg}}$ enforces pixel-wise consistency during sampling, ensuring geometry matches supervision from the outset. \textbf{(2) Class-intrinsic texture quality.} Beyond alignment, our samples exhibit textures and appearances that better capture essential attributes of each semantic category. This is attributable to $\mathcal{L}_{\text{feat}}$, which guides synthesis toward feature distributions measured from real training images, ensuring each sample encodes dataset-level representativeness rather than arbitrary visual diversity. \textbf{(3) Training utility.} The combination of precise alignment and class-representative appearance makes our synthesized samples significantly more \textit{informative}—each image carries clearer supervision signals and more canonical exemplars of its constituent classes. In the distillation regime where every sample must compensate for 100:1 compression, this informativeness directly translates to improved downstream performance, as evidenced by consistent 2.5--4.8 mIoU gains across all architectures.

\textbf{Discussion.} This comparison clarifies the distinction between mask-conditioned synthesis as a \textit{mechanism} versus its \textit{objective}. FreeMask and similar GDA methods optimize for diversity and coverage under the assumption that weak samples are harmless when training includes the full real dataset. Our GDD setting inverts this: the synthesized set must \textit{replace} real data, requiring each sample to be informationally dense and aligned with dataset-level statistics. The results validate that simply applying a GDA pipeline to extreme compression ratios is insufficient—distillation requires objectives that explicitly target representativeness and alignment, which our two-stage design addresses through balanced selection and feature-matching guidance.

\subsection{Class-Wise Performance Analysis}
\label{appendix:Class-Wise Performance Analysis}

\textbf{Motivation.} Aggregate mIoU can obscure performance on categories that matter most for dataset distillation: rare classes nearly absent from random subsets, and difficult classes that are intrinsically hard to synthesize and segment. To provide fine-grained insights, we conduct a class-wise analysis on two groups of five classes per dataset: \textbf{rare}, the lowest-frequency classes in the original training set, and \textbf{difficult}, classes that are \textit{not} rare but intrinsically challenging due to small size, thin structure, or visual ambiguity. This separation disentangles the failure modes of scarcity from those of learnability.

\textbf{Experimental Setup.} We select the 5 lowest-frequency classes as the rare group for each dataset. For the difficult group, we manually identify 5 classes that appear with moderate frequency but pose inherent challenges: thin structures (\textit{poles}, \textit{streetlights}), small objects (\textit{hair drier}), and ambiguous categories (\textit{flags}). The reference baseline is Uniform, the strongest coreset method that also enforces class balance, isolating what Stage II synthesis adds beyond selection. Experiments follow the main paper's configuration: Mask2Former (Swin-T), 2\% distillation on ADE20K and 1\% on COCO-Stuff, averaged over 5 runs. We report per-class IoU, original and distilled class frequencies, and group means. Per-class IoU variance is omitted for brevity (all $\leq$0.7).

\begin{table}[!t]
\centering
\caption{Class-wise performance on ADE20K at 2\% compression (mIoU \%). Freq. indicates the fraction of images containing each class in the original vs. distilled set.}
\label{tab:classwise_ade20k}
\begin{tabular}{l|l|cc|cc}
\toprule
Group & Class (Index) & Freq. (orig.) & Freq. (ours) & Uniform & \textbf{Ours} \\
\midrule
\multirow{5}{*}{\textit{Rare}}
& escalator (97)    & 0.0025\% & 0.38\% & 6.30  & \textbf{18.59} \\
& barrel (112)      & 0.0030\% & 0.38\% & 1.29  & \textbf{18.71} \\
& ship (104)        & 0.0032\% & 0.38\% & 5.98  & \textbf{16.64} \\
& lake (129)        & 0.0032\% & 0.38\% & 7.48  & \textbf{15.10} \\
& conveyer belt (106)& 0.0035\% & 0.34\% & 11.33 & \textbf{30.13} \\
\cmidrule(lr){2-6}
& \textbf{Rare mean} & 0.0031\% & 0.37\% & 6.48 & \textbf{19.83} \\
\midrule
\multirow{5}{*}{\textit{Difficult}}
& pole (94)         & 0.60\% & 0.47\% & 9.75  & \textbf{11.18} \\
& streetlight (88)  & 1.20\% & 0.59\% & 11.64 & \textbf{19.65} \\
& traffic light (137)& 0.30\% & 0.34\% & 9.73  & \textbf{12.36} \\
& sconce (135)      & 0.62\% & 0.59\% & 10.79 & \textbf{17.48} \\
& flag (150)        & 0.26\% & 0.42\% & 0.00  & \textbf{16.98} \\
\cmidrule(lr){2-6}
& \textbf{Difficult mean} & 0.60\% & 0.48\% & 8.38 & \textbf{15.53} \\
\midrule
\multicolumn{2}{c|}{\textbf{All-class mIoU}} & -- & -- & 23.19 & \textbf{26.13} \\
\bottomrule
\end{tabular}
\end{table}

\begin{table}[!t]
\centering
\caption{Class-wise performance on COCO-Stuff at 1\% compression (mIoU \%). Freq. indicates the fraction of images containing each class in the original vs. distilled set.}
\label{tab:classwise_coco}
\begin{tabular}{l|l|cc|cc}
\toprule
Group & Class (Index) & Freq. (orig.) & Freq. (ours) & Uniform & \textbf{Ours} \\
\midrule
\multirow{5}{*}{\textit{Rare}}
& waterdrops (168)  & 0.0012\% & 0.35\% & 2.82  & \textbf{9.70} \\
& hair drier (79)   & 0.0020\% & 0.35\% & 1.40  & \textbf{12.32} \\
& toaster (71)      & 0.0023\% & 0.35\% & 22.89 & \textbf{30.50} \\
& moss (123)        & 0.0027\% & 0.35\% & 7.80  & \textbf{14.10} \\
& ceiling-tile (92) & 0.0036\% & 0.35\% & 11.96 & \textbf{19.27} \\
\cmidrule(lr){2-6}
& \textbf{Rare mean} & 0.0024\% & 0.35\% & 9.34 & \textbf{17.18} \\
\midrule
\multirow{5}{*}{\textit{Difficult}}
& toothbrush (80)   & 0.09\% & 0.35\% & 10.71 & \textbf{15.18} \\
& scissors (77)     & 0.10\% & 0.35\% & 21.29 & \textbf{44.06} \\
& spoon (45)        & 0.33\% & 0.38\% & 2.76  & \textbf{9.54} \\
& knife (44)        & 0.41\% & 0.54\% & 4.85  & \textbf{10.28} \\
& fork (43)         & 0.33\% & 0.37\% & 7.88  & \textbf{19.73} \\
\cmidrule(lr){2-6}
& \textbf{Difficult mean} & 0.25\% & 0.40\% & 9.45 & \textbf{19.76} \\
\midrule
\multicolumn{2}{c|}{\textbf{All-class mIoU}} & -- & -- & 28.96 & \textbf{31.19} \\
\bottomrule
\end{tabular}
\end{table}

\begin{figure}[!h]
\centering
\includegraphics[width=\textwidth]{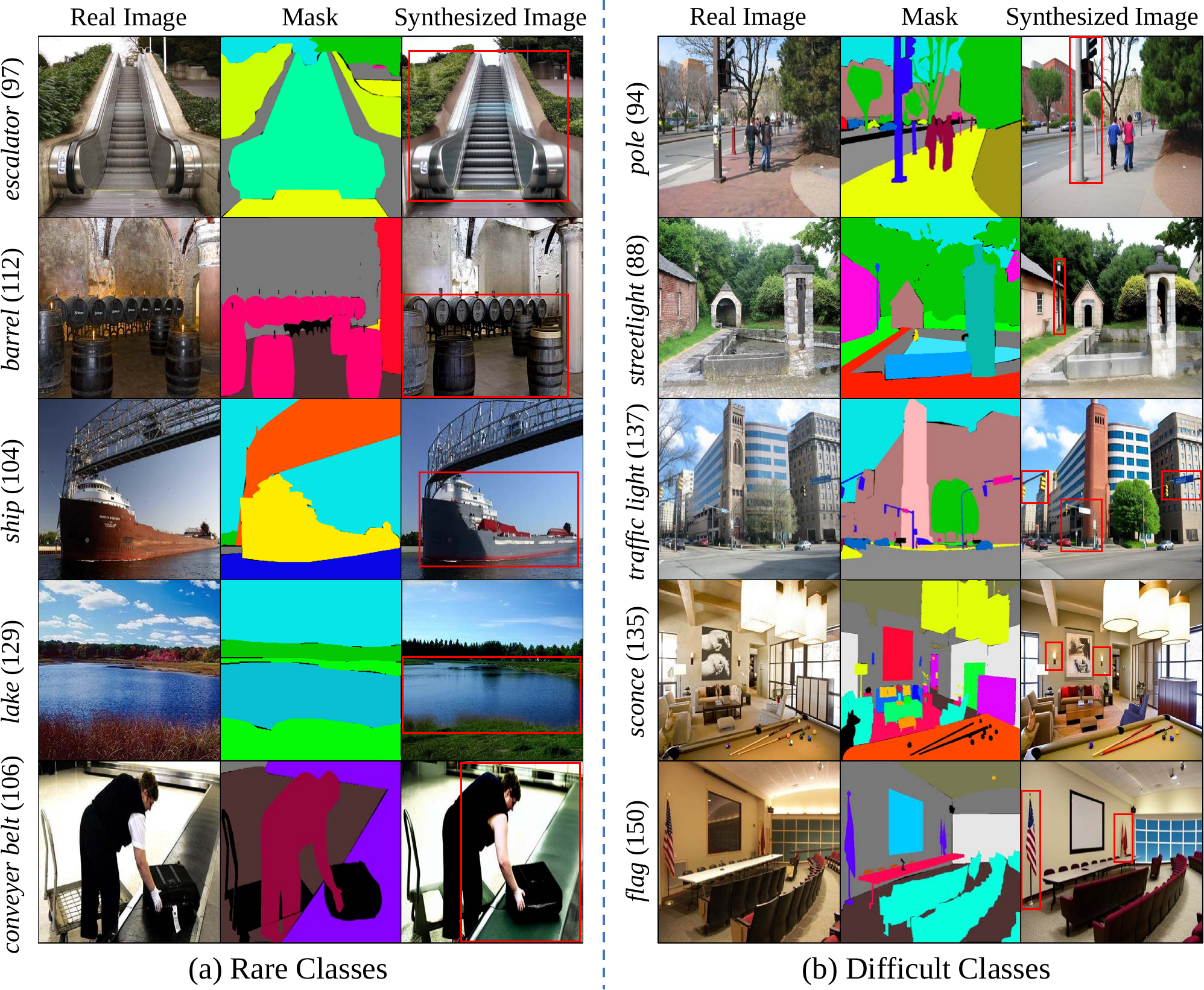}
\caption{Qualitative examples for rare and difficult classes on ADE20K. For each class, we show the real image, segmentation mask, and our synthesized image. Stage I lifts rare class representation to ~0.35--0.40\%, while Stage II generates clean, canonical instances that enhance learnability even for thin structures.}
\label{fig:classwise_ade20k}
\end{figure}

\begin{figure}[!h]
\centering
\includegraphics[width=\textwidth]{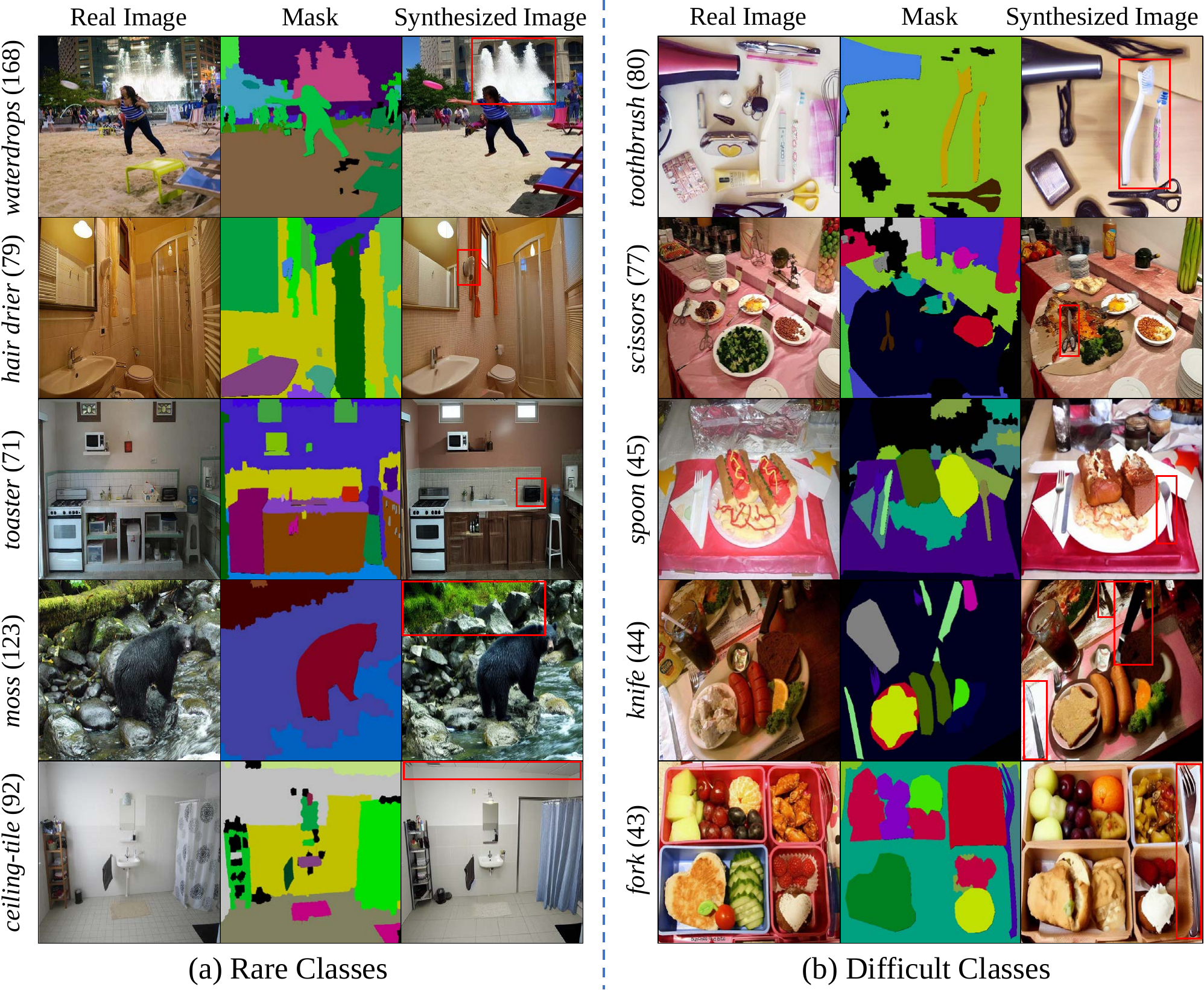}
\caption{Qualitative examples for rare and difficult classes on COCO-Stuff. For each class, we show the real image, segmentation mask, and our synthesized image. Our method generates visually distinct and class-discriminative instances even for small, ambiguous objects and uncommon stuff categories.}
\label{fig:classwise_coco}
\end{figure}

\textbf{Results and Analysis.} Four key patterns emerge from~\Cref{tab:classwise_ade20k,tab:classwise_coco} and~\Cref{fig:classwise_ade20k,fig:classwise_coco}: \textbf{(1) Gains concentrate where they matter most.} On ADE20K, rare classes gain +13.35 IoU (6.48 $\to$ 19.83) and difficult classes +7.15 IoU (8.38 $\to$ 15.53), compared to an all-class gain of only +2.94—representing $4.5\times$ and $2.4\times$ amplification, respectively. On COCO-Stuff, the pattern is even stronger: +7.84 IoU for rare classes and +10.31 IoU for difficult classes, versus +2.23 overall ($3.5\times$ and $4.6\times$ amplification). This demonstrates that aggregate mIoU significantly understates what our method accomplishes for the categories that matter most in dataset distillation—those that are scarce or inherently challenging. \textbf{(2) Two mechanisms map onto two stages.} For rare classes, Stage I is the primary driver: representation increases by ~120$\times$ on ADE20K (0.0031\% $\to$ 0.37\%) and ~146$\times$ on COCO-Stuff (0.0024\% $\to$ 0.35\%), both converging to the ~0.35--0.40\% coverage floor that balanced selection guarantees at these compression ratios. This direct exposure lift explains the IoU jumps. For difficult classes, however, representation is \textit{not} the explanation. On ADE20K, their frequency actually \textit{decreases} (0.60\% $\to$ 0.48\%), with \textit{streetlight} dropping from 1.20\% to 0.59\%, yet IoU still gains +7.15. On COCO-Stuff, the modest increase (0.25\% $\to$ 0.40\%) cannot account for the +10.31 gain. What changes is the \textit{content} of each sample—Stage II synthesizes cleaner, more canonical instances than the cluttered real images they replace, as visualized in~\Cref{fig:classwise_ade20k,fig:classwise_coco}. Thus, Stage I fixes scarcity through balanced selection, while Stage II fixes learnability through informative synthesis. \textbf{(3) Balance over real images is insufficient.} Uniform also enforces class balance, yet leaves several categories nearly unlearnable: \textit{flag} at 0.00 IoU, \textit{barrel} at 1.29, \textit{hair drier} at 1.40, \textit{spoon} at 2.76, and \textit{waterdrops} at 2.82. Our distilled set lifts these to 16.98, 18.71, 12.32, 9.54, and 9.70, respectively—gains of +17.42 to +6.88 IoU from near-zero baselines. This reveals a critical insight: at extreme compression, a few real images where a rare class appears small and occluded provide little learnable signal. Simply balancing class coverage is not enough—what matters is the \textit{informativeness} of each sample. Our Stage II guidance ensures that every synthesized image presents clear, class-discriminative features aligned with dataset-level statistics, making each sample carry more training utility than its real counterpart. \textbf{(4) Honest limits and concrete remedies.} Absolute IoU remains modest for the thinnest structures: \textit{pole} gains only +1.43 (9.75 $\to$ 11.18), and categories like \textit{spoon} (9.54), \textit{waterdrops} (9.70), and \textit{toothbrush} (15.18) remain relatively low despite significant improvements. This limitation traces to the layout-to-image generator's fidelity on near-pixel-wide geometry rather than a flaw in our objective design. Since Stage I supplies \textit{real} masks with exact boundaries, what limits these classes is the generator's rendering competence—consistent with our findings in the FreeMask comparison, where a weaker prior costs appearance realism while geometry and supervision survive. Stronger diffusion backbones (e.g., higher-resolution generators) or boundary-aware guidance terms are concrete remedies that future work can explore. Importantly, even with this bottleneck, every single class improves over Uniform, and substantial gains (+8.01 for \textit{streetlight}, +22.77 for \textit{scissors}) demonstrate that the framework is not fundamentally limited to common or easy categories.

\subsection{Comparison with Advanced Coreset Selection Methods}
\label{appendix:Comparison with Advanced Coreset Selection Methods}

\textbf{Motivation.} The baselines in the main paper (Random, K-Center, Herding) represent foundational coreset principles but do not reflect recent advances. To rigorously validate our class-balanced selection strategy, we compare against three state-of-the-art coreset methods that embody modern design principles: \textbf{Coverage-centric Coreset Selection (CCS)}~\cite{CCS} optimizes coverage over feature space with high pruning rates; \textbf{D2Pruning}~\cite{D2Pruning} balances diversity and sample difficulty via message passing; and \textbf{Structural-Entropy-based Sample Selection (SES)}~\cite{SES} scores samples by their structural entropy contribution. These methods represent the frontier of coreset research in classification, and comparing against them tests whether class-level coverage is genuinely necessary or whether advanced diversity-based selection suffices.


\textbf{Experimental Setup.} We evaluate all three methods on ADE20K using Mask2Former (Swin-S) across compression ratios of 0.5\%, 1\%, and 2\%. For fair comparison, we use each method's official implementation with default hyperparameters. We adapt CCS, D2Pruning, and SES to semantic segmentation using image-level difficulty estimates derived from a Mask2Former model trained on the full ADE20K training set. SES and D2Pruning additionally construct image-similarity graphs from global Swin-S features, extracted following the same protocol as all operate on global image-level features extracted from a pretrained Swin-S backbone (same as used for K-Center and Herding). All methods select real images, which are then used for segmentation training following the protocol in~\Cref{tab:ade20k_distill}. Results are averaged over 5 runs.

\begin{table}[!t]
\centering
\caption{Comparison with advanced coreset selection methods on ADE20K using Mask2Former (Swin-S), mIoU (\%).}
\label{tab:advanced_coreset}
\begin{tabular}{l|ccc}
\toprule
Method & 0.5\% & 1\% & 2\% \\
\midrule
Random & 10.98{\scriptsize $\pm$0.5} & 15.65{\scriptsize $\pm$0.5} & 21.70{\scriptsize $\pm$0.4} \\
K-Center & 10.52{\scriptsize $\pm$0.4} & 13.76{\scriptsize $\pm$0.4} & 17.67{\scriptsize $\pm$0.3} \\
Herding & 12.47{\scriptsize $\pm$0.4} & 16.32{\scriptsize $\pm$0.3} & 22.81{\scriptsize $\pm$0.4} \\
Uniform & 15.04{\scriptsize $\pm$0.2} & 21.42{\scriptsize $\pm$0.3} & 27.59{\scriptsize $\pm$0.3} \\
\midrule
CCS & 8.25{\scriptsize $\pm$0.4} & 10.67{\scriptsize $\pm$0.4} & 17.61{\scriptsize $\pm$0.3} \\
SES & 13.64{\scriptsize $\pm$0.3} & 17.81{\scriptsize $\pm$0.3} & 24.33{\scriptsize $\pm$0.2} \\
D2Pruning & 14.51{\scriptsize $\pm$0.5} & 19.28{\scriptsize $\pm$0.4} & 24.50{\scriptsize $\pm$0.3} \\
\midrule
\textbf{Ours} & \textbf{19.45{\scriptsize $\pm$0.5}} & \textbf{24.99{\scriptsize $\pm$0.5}} & \textbf{30.28{\scriptsize $\pm$0.5}} \\
\midrule
\textit{Full Dataset} & \textit{51.22{\scriptsize $\pm$0.1}} & \textit{51.22{\scriptsize $\pm$0.1}} & \textit{51.22{\scriptsize $\pm$0.1}} \\
\bottomrule
\end{tabular}
\end{table}

\textbf{Results and Analysis.} Three key findings emerge from~\Cref{tab:advanced_coreset}: \textbf{(1) Advanced methods are stronger, and our margin holds.} D2Pruning and SES substantially outperform the foundational baselines, confirming they represent the state of the art. D2Pruning gains +1.04 to +2.04 over Herding and +1.53 to +3.63 over Random, while SES gains +1.17 to +1.49 over Herding. Despite this, \MethodName\ maintains significant advantages: +4.94 mIoU over D2Pruning at 0.5\%, +5.71 at 1\%, and +5.78 at 2\%. Crucially, the margin \textit{grows} as the budget tightens—the regime dataset distillation targets—demonstrating that our approach is not merely competitive but addresses a fundamentally different bottleneck. \textbf{(2) Class balance is the right principle for dense prediction under extreme compression.} Remarkably, Uniform—the simplest class-balanced baseline with no learned scoring—outperforms all three advanced methods at every compression ratio. The gap widens from +0.53 mIoU over D2Pruning at 0.5\% to +2.14 at 1\% and +3.09 at 2\%. This is not a defect of CCS, SES, or D2Pruning but a consequence of the task shift from classification to dense prediction. All three methods assign a scalar score to each whole image, which is well-defined when one image carries one label. In semantic segmentation, a multi-class scene contains 10--30 overlapping categories, and a scalar score necessarily averages away the tail signal—an image containing one rare class among twenty common ones becomes indistinguishable from an ordinary scene. With no mechanism enforcing per-class coverage, ADE20K's severe imbalance (Imbalance Factor = 282) survives selection, leaving tail categories nearly absent from the distilled subset. \textbf{(3) Synthesis amplifies the advantage.} While Uniform beats D2Pruning by +2.14 to +3.09 mIoU, our full pipeline extends this to +5.71 to +5.78, demonstrating that Stage II synthesis compounds the benefit of balanced selection. The selected masks provide diverse, class-representative layouts, and guided diffusion refines them into canonical, informative training samples. This two-stage design—class-aware selection followed by feature-matching synthesis—proves essential for segmentation distillation, where both coverage and sample quality matter more than in classification.

\textbf{Discussion.} This comparison clarifies a broader principle: coreset methods designed for classification excel at capturing global diversity and sample difficulty, but these criteria are \textit{class-agnostic}. For tasks with severe long-tail distributions—where performance hinges on tail coverage—scalar image scoring cannot distinguish a rare-class-bearing image from a common one. Our class-balanced selection explicitly tracks per-class coverage, ensuring no category is neglected. Combined with diffusion-guided synthesis, this yields a distilled set that is both balanced and informative, outperforming even the most advanced diversity-based methods by a substantial margin.

\section{Limitations and Future Works}
\label{appendix:Limitations and Future Works}

\begin{figure}[!t]
    \centering
    \includegraphics[width=\textwidth]{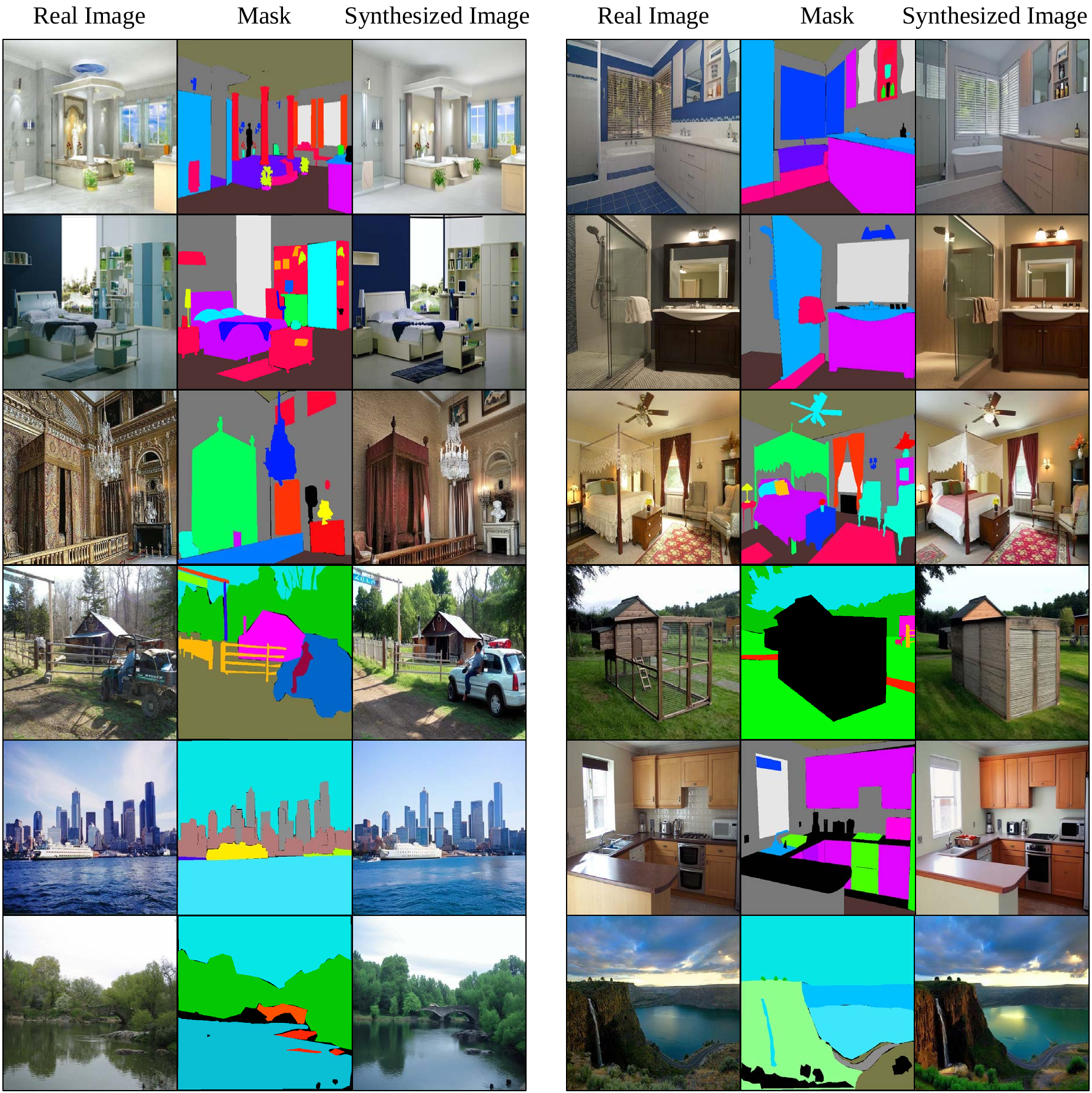}
    \caption{Additional qualitative results on ADE20K. Each example shows the real image, segmentation mask, and synthesized image generated by \MethodName, demonstrating strong layout alignment and clear semantic structures.}
    \label{fig:ade_more_samples}
\end{figure}

\begin{figure}[!t]
    \centering
    \includegraphics[width=\textwidth]{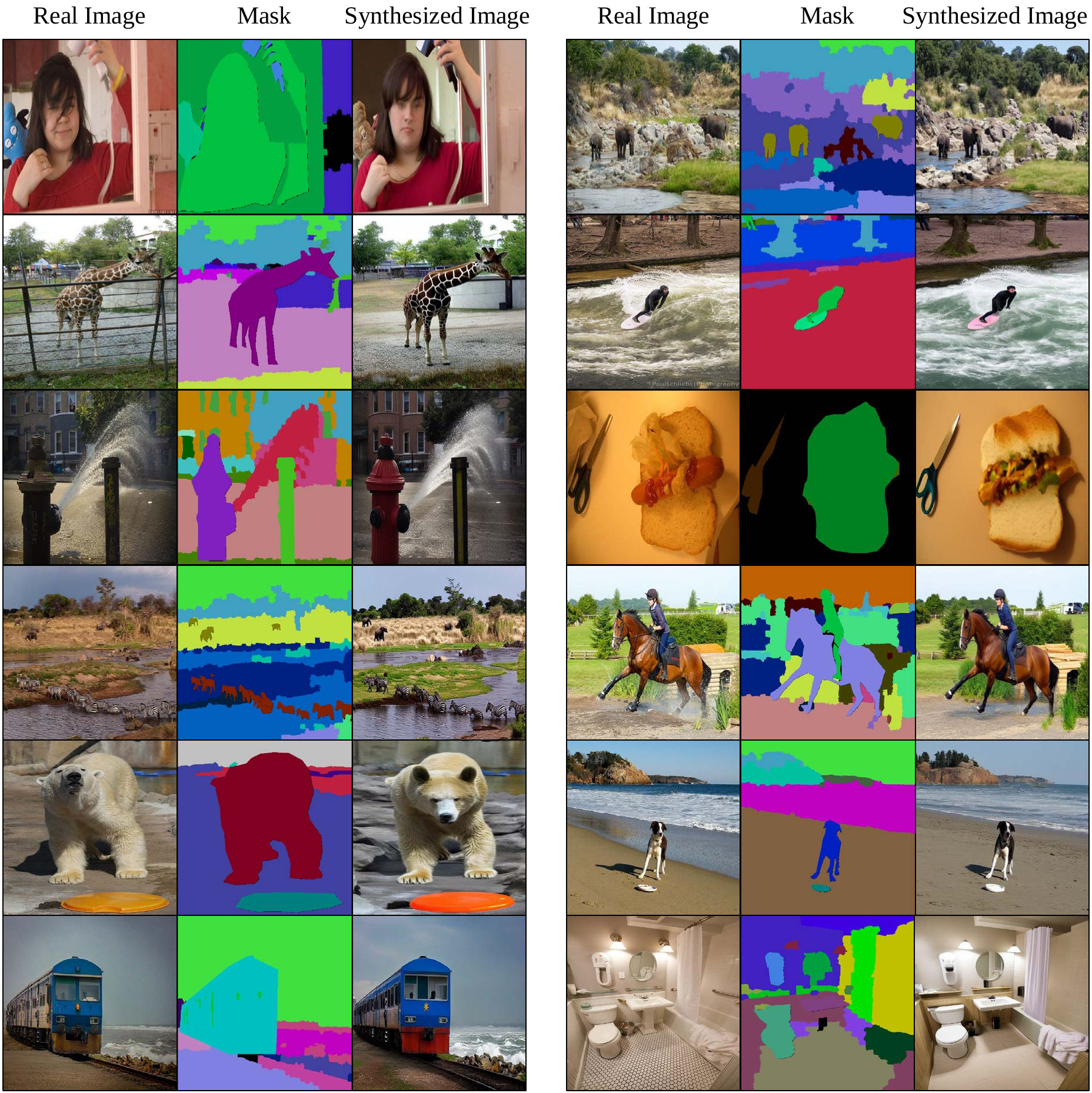}
    \caption{Additional qualitative results on COCO-Stuff. The synthesized images exhibit consistent alignment with input masks and emphasize class-discriminative features across diverse and complex scenes.}
    \label{fig:coco_more_samples}
\end{figure}

Despite the strong empirical performance, \MethodName\ has several limitations that suggest promising directions for future research.
\textbf{First}, our framework relies on a layout-to-image diffusion model pretrained on the target dataset. When such a model is unavailable, a domain gap may arise, leading to degraded generation quality and reduced effectiveness of the distilled data. In future work, this limitation could be alleviated by leveraging more general-purpose diffusion models with stronger cross-domain generalization or by introducing lightweight adaptation techniques (e.g., parameter-efficient fine-tuning) to quickly align pretrained models with new target domains. \textbf{Second}, the proposed guided sampling strategy requires a pretrained segmentation model as the guidance network. For datasets lacking publicly available checkpoints, training such a model from scratch introduces additional computational overhead. A potential direction is to reduce this dependency by exploring self-supervised or weakly supervised alternatives for guidance, or by distilling transferable segmentation priors from large-scale foundation models to new domains with minimal supervision. \textbf{Finally}, our method directly selects masks from the original dataset, which may not be optimal under extreme compression ratios. The selected layouts are constrained by existing data distributions and cannot fully explore the space of informative semantic configurations. A promising direction is to move beyond selection-based strategies and instead jointly synthesize images and layouts~\cite{JoDiffusion}. Such a layout–image co-generation could enable more flexible and diverse semantic compositions, potentially leading to more compact and informative distilled datasets.

\section{Broader Impacts}
\label{appendix:Broader Impacts}

This work introduces a diffusion-based framework for dataset distillation in semantic segmentation, enabling efficient training with significantly reduced data. By lowering the data and computational requirements, our method can facilitate research and deployment in resource-constrained settings, such as edge devices or domains where large-scale annotation is costly (e.g., medical or environmental applications). However, as the method relies on generative models, it may inherit biases present in the original dataset, potentially amplifying class imbalance or under-representation in downstream tasks. In addition, synthesized data may introduce artifacts or unrealistic patterns that could affect model robustness if not properly controlled. There is also a risk that synthetic datasets could be misused to obscure data provenance or bypass data governance policies. Future work should explore bias mitigation, improved controllability of generation, and mechanisms for ensuring transparency and traceability of synthetic data. Overall, we believe this line of research has the potential to make data-efficient learning more accessible while highlighting the importance of responsible use of generative models.

\section{More Visualizations}
\label{appendix:more_visualizations}

\Cref{fig:ade_more_samples} and \Cref{fig:coco_more_samples} present additional qualitative results on ADE20K and COCO-Stuff, respectively. Each example includes the real image, the corresponding segmentation mask, and the synthesized image generated by \MethodName. Consistent with the observations in the main paper, the synthesized images exhibit strong spatial alignment with the input masks across diverse scenes. Object boundaries and spatial structures are well preserved, indicating effective pixel-level consistency during generation. Moreover, the generated samples highlight class-discriminative features while suppressing irrelevant variations, resulting in cleaner and more canonical representations compared to real images. Even in complex layouts with multiple co-occurring objects, the synthesized images maintain clear foreground structures and semantic coherence. These results further demonstrate that the proposed diffusion-guided framework produces structurally faithful and semantically informative samples, which are well-suited for training segmentation models.

\section{Large Language Model Usage}
\label{appendix:llm_usage}

We use ChatGPT (GPT-5, by OpenAI) as a writing assistant to polish the language of this paper. Specifically, it was employed to improve clarity, grammar, and style in certain sections, but it did not contribute to research ideation, technical content, experiments, or analysis. All scientific contributions, methods, and conclusions are solely the work of the authors. The authors take full responsibility for the accuracy and integrity of the paper's contents.


\newpage
\input{checklist.tex}

\end{document}

%% file: checklist.tex
\section*{NeurIPS Paper Checklist}

\begin{enumerate}

\item {\bf Claims}
    \item[] Question: Do the main claims made in the abstract and introduction accurately reflect the paper's contributions and scope?
    \item[] Answer: \answerYes{} 
    \item[] Justification: Our abstract and introduction accurately reflect the paper’s contributions to diffusion-guided dataset distillation for semantic segmentation.
    \item[] Guidelines:
    \begin{itemize}
        \item The answer \answerNA{} means that the abstract and introduction do not include the claims made in the paper.
        \item The abstract and/or introduction should clearly state the claims made, including the contributions made in the paper and important assumptions and limitations. A \answerNo{} or \answerNA{} answer to this question will not be perceived well by the reviewers. 
        \item The claims made should match theoretical and experimental results, and reflect how much the results can be expected to generalize to other settings. 
        \item It is fine to include aspirational goals as motivation as long as it is clear that these goals are not attained by the paper. 
    \end{itemize}

\item {\bf Limitations}
    \item[] Question: Does the paper discuss the limitations of the work performed by the authors?
    \item[] Answer: \answerYes{} 
    \item[] Justification: Yes, we discuss this in Appendix~\ref{appendix:Limitations and Future Works}.
    \item[] Guidelines:
    \begin{itemize}
        \item The answer \answerNA{} means that the paper has no limitation while the answer \answerNo{} means that the paper has limitations, but those are not discussed in the paper. 
        \item The authors are encouraged to create a separate ``Limitations'' section in their paper.
        \item The paper should point out any strong assumptions and how robust the results are to violations of these assumptions (e.g., independence assumptions, noiseless settings, model well-specification, asymptotic approximations only holding locally). The authors should reflect on how these assumptions might be violated in practice and what the implications would be.
        \item The authors should reflect on the scope of the claims made, e.g., if the approach was only tested on a few datasets or with a few runs. In general, empirical results often depend on implicit assumptions, which should be articulated.
        \item The authors should reflect on the factors that influence the performance of the approach. For example, a facial recognition algorithm may perform poorly when image resolution is low or images are taken in low lighting. Or a speech-to-text system might not be used reliably to provide closed captions for online lectures because it fails to handle technical jargon.
        \item The authors should discuss the computational efficiency of the proposed algorithms and how they scale with dataset size.
        \item If applicable, the authors should discuss possible limitations of their approach to address problems of privacy and fairness.
        \item While the authors might fear that complete honesty about limitations might be used by reviewers as grounds for rejection, a worse outcome might be that reviewers discover limitations that aren't acknowledged in the paper. The authors should use their best judgment and recognize that individual actions in favor of transparency play an important role in developing norms that preserve the integrity of the community. Reviewers will be specifically instructed to not penalize honesty concerning limitations.
    \end{itemize}

\item {\bf Theory assumptions and proofs}
    \item[] Question: For each theoretical result, does the paper provide the full set of assumptions and a complete (and correct) proof?
    \item[] Answer: \answerNA{} 
    \item[] Justification: This paper does not include theorems or lemmas.
    \item[] Guidelines:
    \begin{itemize}
        \item The answer \answerNA{} means that the paper does not include theoretical results. 
        \item All the theorems, formulas, and proofs in the paper should be numbered and cross-referenced.
        \item All assumptions should be clearly stated or referenced in the statement of any theorems.
        \item The proofs can either appear in the main paper or the supplemental material, but if they appear in the supplemental material, the authors are encouraged to provide a short proof sketch to provide intuition. 
        \item Inversely, any informal proof provided in the core of the paper should be complemented by formal proofs provided in appendix or supplemental material.
        \item Theorems and Lemmas that the proof relies upon should be properly referenced. 
    \end{itemize}

    \item {\bf Experimental result reproducibility}
    \item[] Question: Does the paper fully disclose all the information needed to reproduce the main experimental results of the paper to the extent that it affects the main claims and/or conclusions of the paper (regardless of whether the code and data are provided or not)?
    \item[] Answer: \answerYes{} 
    \item[] Justification: Yes, we provide sufficient implementation details in~\Cref{sec:Experiment Setup} and Appendix~\ref{appendix:Segmentation Model Training}.
    \item[] Guidelines:
    \begin{itemize}
        \item The answer \answerNA{} means that the paper does not include experiments.
        \item If the paper includes experiments, a \answerNo{} answer to this question will not be perceived well by the reviewers: Making the paper reproducible is important, regardless of whether the code and data are provided or not.
        \item If the contribution is a dataset and\slash or model, the authors should describe the steps taken to make their results reproducible or verifiable. 
        \item Depending on the contribution, reproducibility can be accomplished in various ways. For example, if the contribution is a novel architecture, describing the architecture fully might suffice, or if the contribution is a specific model and empirical evaluation, it may be necessary to either make it possible for others to replicate the model with the same dataset, or provide access to the model. In general. releasing code and data is often one good way to accomplish this, but reproducibility can also be provided via detailed instructions for how to replicate the results, access to a hosted model (e.g., in the case of a large language model), releasing of a model checkpoint, or other means that are appropriate to the research performed.
        \item While NeurIPS does not require releasing code, the conference does require all submissions to provide some reasonable avenue for reproducibility, which may depend on the nature of the contribution. For example
        \begin{enumerate}
            \item If the contribution is primarily a new algorithm, the paper should make it clear how to reproduce that algorithm.
            \item If the contribution is primarily a new model architecture, the paper should describe the architecture clearly and fully.
            \item If the contribution is a new model (e.g., a large language model), then there should either be a way to access this model for reproducing the results or a way to reproduce the model (e.g., with an open-source dataset or instructions for how to construct the dataset).
            \item We recognize that reproducibility may be tricky in some cases, in which case authors are welcome to describe the particular way they provide for reproducibility. In the case of closed-source models, it may be that access to the model is limited in some way (e.g., to registered users), but it should be possible for other researchers to have some path to reproducing or verifying the results.
        \end{enumerate}
    \end{itemize}

\item {\bf Open access to data and code}
    \item[] Question: Does the paper provide open access to the data and code, with sufficient instructions to faithfully reproduce the main experimental results, as described in supplemental material?
    \item[] Answer: \answerYes{} 
    \item[] Justification: We provide code in the supplementary materials and provide a GitHub link in the Abstract and Appendix~\ref{appendix:Segmentation Model Training}.
    \item[] Guidelines:
    \begin{itemize}
        \item The answer \answerNA{} means that paper does not include experiments requiring code.
        \item Please see the NeurIPS code and data submission guidelines (\url{https://neurips.cc/public/guides/CodeSubmissionPolicy}) for more details.
        \item While we encourage the release of code and data, we understand that this might not be possible, so \answerNo{} is an acceptable answer. Papers cannot be rejected simply for not including code, unless this is central to the contribution (e.g., for a new open-source benchmark).
        \item The instructions should contain the exact command and environment needed to run to reproduce the results. See the NeurIPS code and data submission guidelines (\url{https://neurips.cc/public/guides/CodeSubmissionPolicy}) for more details.
        \item The authors should provide instructions on data access and preparation, including how to access the raw data, preprocessed data, intermediate data, and generated data, etc.
        \item The authors should provide scripts to reproduce all experimental results for the new proposed method and baselines. If only a subset of experiments are reproducible, they should state which ones are omitted from the script and why.
        \item At submission time, to preserve anonymity, the authors should release anonymized versions (if applicable).
        \item Providing as much information as possible in supplemental material (appended to the paper) is recommended, but including URLs to data and code is permitted.
    \end{itemize}

\item {\bf Experimental setting/details}
    \item[] Question: Does the paper specify all the training and test details (e.g., data splits, hyperparameters, how they were chosen, type of optimizer) necessary to understand the results?
    \item[] Answer: \answerYes{} 
    \item[] Justification: Experimental setting is described in~\Cref{sec:Experiment Setup} and Appendix~\ref{appendix:Segmentation Model Training}.
    \item[] Guidelines:
    \begin{itemize}
        \item The answer \answerNA{} means that the paper does not include experiments.
        \item The experimental setting should be presented in the core of the paper to a level of detail that is necessary to appreciate the results and make sense of them.
        \item The full details can be provided either with the code, in appendix, or as supplemental material.
    \end{itemize}

\item {\bf Experiment statistical significance}
    \item[] Question: Does the paper report error bars suitably and correctly defined or other appropriate information about the statistical significance of the experiments?
    \item[] Answer: \answerYes{} 
    \item[] Justification: We report error bars in~\Cref{tab:ade20k_distill,tab:coco_distill,tab:ablation_components,tab:cross_arch}.
    \item[] Guidelines:
    \begin{itemize}
        \item The answer \answerNA{} means that the paper does not include experiments.
        \item The authors should answer \answerYes{} if the results are accompanied by error bars, confidence intervals, or statistical significance tests, at least for the experiments that support the main claims of the paper.
        \item The factors of variability that the error bars are capturing should be clearly stated (for example, train/test split, initialization, random drawing of some parameter, or overall run with given experimental conditions).
        \item The method for calculating the error bars should be explained (closed form formula, call to a library function, bootstrap, etc.)
        \item The assumptions made should be given (e.g., Normally distributed errors).
        \item It should be clear whether the error bar is the standard deviation or the standard error of the mean.
        \item It is OK to report 1-sigma error bars, but one should state it. The authors should preferably report a 2-sigma error bar than state that they have a 96\% CI, if the hypothesis of Normality of errors is not verified.
        \item For asymmetric distributions, the authors should be careful not to show in tables or figures symmetric error bars that would yield results that are out of range (e.g., negative error rates).
        \item If error bars are reported in tables or plots, the authors should explain in the text how they were calculated and reference the corresponding figures or tables in the text.
    \end{itemize}

\item {\bf Experiments compute resources}
    \item[] Question: For each experiment, does the paper provide sufficient information on the computer resources (type of compute workers, memory, time of execution) needed to reproduce the experiments?
    \item[] Answer: \answerYes{} 
    \item[] Justification: We provide the computer resources in~\Cref{sec:Experiment Setup} and Appendix~\ref{appendix:Segmentation Model Training}.
    \item[] Guidelines:
    \begin{itemize}
        \item The answer \answerNA{} means that the paper does not include experiments.
        \item The paper should indicate the type of compute workers CPU or GPU, internal cluster, or cloud provider, including relevant memory and storage.
        \item The paper should provide the amount of compute required for each of the individual experimental runs as well as estimate the total compute. 
        \item The paper should disclose whether the full research project required more compute than the experiments reported in the paper (e.g., preliminary or failed experiments that didn't make it into the paper). 
    \end{itemize}
    
\item {\bf Code of ethics}
    \item[] Question: Does the research conducted in the paper conform, in every respect, with the NeurIPS Code of Ethics \url{https://neurips.cc/public/EthicsGuidelines}?
    \item[] Answer: \answerYes{} 
    \item[] Justification: Our research conform with the NeurIPS Code of Ethics.
    \item[] Guidelines:
    \begin{itemize}
        \item The answer \answerNA{} means that the authors have not reviewed the NeurIPS Code of Ethics.
        \item If the authors answer \answerNo, they should explain the special circumstances that require a deviation from the Code of Ethics.
        \item The authors should make sure to preserve anonymity (e.g., if there is a special consideration due to laws or regulations in their jurisdiction).
    \end{itemize}

\item {\bf Broader impacts}
    \item[] Question: Does the paper discuss both potential positive societal impacts and negative societal impacts of the work performed?
    \item[] Answer: \answerYes{} 
    \item[] Justification: We discuss the broader impact in Appendix~\ref{appendix:Broader Impacts}.
    \item[] Guidelines:
    \begin{itemize}
        \item The answer \answerNA{} means that there is no societal impact of the work performed.
        \item If the authors answer \answerNA{} or \answerNo, they should explain why their work has no societal impact or why the paper does not address societal impact.
        \item Examples of negative societal impacts include potential malicious or unintended uses (e.g., disinformation, generating fake profiles, surveillance), fairness considerations (e.g., deployment of technologies that could make decisions that unfairly impact specific groups), privacy considerations, and security considerations.
        \item The conference expects that many papers will be foundational research and not tied to particular applications, let alone deployments. However, if there is a direct path to any negative applications, the authors should point it out. For example, it is legitimate to point out that an improvement in the quality of generative models could be used to generate Deepfakes for disinformation. On the other hand, it is not needed to point out that a generic algorithm for optimizing neural networks could enable people to train models that generate Deepfakes faster.
        \item The authors should consider possible harms that could arise when the technology is being used as intended and functioning correctly, harms that could arise when the technology is being used as intended but gives incorrect results, and harms following from (intentional or unintentional) misuse of the technology.
        \item If there are negative societal impacts, the authors could also discuss possible mitigation strategies (e.g., gated release of models, providing defenses in addition to attacks, mechanisms for monitoring misuse, mechanisms to monitor how a system learns from feedback over time, improving the efficiency and accessibility of ML).
    \end{itemize}
    
\item {\bf Safeguards}
    \item[] Question: Does the paper describe safeguards that have been put in place for responsible release of data or models that have a high risk for misuse (e.g., pre-trained language models, image generators, or scraped datasets)?
    \item[] Answer: \answerNA{} 
    \item[] Justification: The paper poses no such risks.
    \item[] Guidelines:
    \begin{itemize}
        \item The answer \answerNA{} means that the paper poses no such risks.
        \item Released models that have a high risk for misuse or dual-use should be released with necessary safeguards to allow for controlled use of the model, for example by requiring that users adhere to usage guidelines or restrictions to access the model or implementing safety filters. 
        \item Datasets that have been scraped from the Internet could pose safety risks. The authors should describe how they avoided releasing unsafe images.
        \item We recognize that providing effective safeguards is challenging, and many papers do not require this, but we encourage authors to take this into account and make a best faith effort.
    \end{itemize}

\item {\bf Licenses for existing assets}
    \item[] Question: Are the creators or original owners of assets (e.g., code, data, models), used in the paper, properly credited and are the license and terms of use explicitly mentioned and properly respected?
    \item[] Answer: \answerYes{} 
    \item[] Justification: All assets used in the paper, including datasets, are publicly available. Proper credits are given to the creators or original owners of these datasets where applicable. The licenses and terms of use for these datasets are explicitly mentioned and respected in accordance with their respective guidelines.
    \item[] Guidelines:
    \begin{itemize}
        \item The answer \answerNA{} means that the paper does not use existing assets.
        \item The authors should cite the original paper that produced the code package or dataset.
        \item The authors should state which version of the asset is used and, if possible, include a URL.
        \item The name of the license (e.g., CC-BY 4.0) should be included for each asset.
        \item For scraped data from a particular source (e.g., website), the copyright and terms of service of that source should be provided.
        \item If assets are released, the license, copyright information, and terms of use in the package should be provided. For popular datasets, \url{paperswithcode.com/datasets} has curated licenses for some datasets. Their licensing guide can help determine the license of a dataset.
        \item For existing datasets that are re-packaged, both the original license and the license of the derived asset (if it has changed) should be provided.
        \item If this information is not available online, the authors are encouraged to reach out to the asset's creators.
    \end{itemize}

\item {\bf New assets}
    \item[] Question: Are new assets introduced in the paper well documented and is the documentation provided alongside the assets?
    \item[] Answer: \answerYes{} 
    \item[] Justification: We have attached our code and user instructions in the supplementary materials.
    \item[] Guidelines:
    \begin{itemize}
        \item The answer \answerNA{} means that the paper does not release new assets.
        \item Researchers should communicate the details of the dataset\slash code\slash model as part of their submissions via structured templates. This includes details about training, license, limitations, etc. 
        \item The paper should discuss whether and how consent was obtained from people whose asset is used.
        \item At submission time, remember to anonymize your assets (if applicable). You can either create an anonymized URL or include an anonymized zip file.
    \end{itemize}

\item {\bf Crowdsourcing and research with human subjects}
    \item[] Question: For crowdsourcing experiments and research with human subjects, does the paper include the full text of instructions given to participants and screenshots, if applicable, as well as details about compensation (if any)? 
    \item[] Answer: \answerNA{} 
    \item[] Justification: The paper does not involve crowdsourcing nor research with human subjects.
    \item[] Guidelines:
    \begin{itemize}
        \item The answer \answerNA{} means that the paper does not involve crowdsourcing nor research with human subjects.
        \item Including this information in the supplemental material is fine, but if the main contribution of the paper involves human subjects, then as much detail as possible should be included in the main paper. 
        \item According to the NeurIPS Code of Ethics, workers involved in data collection, curation, or other labor should be paid at least the minimum wage in the country of the data collector. 
    \end{itemize}

\item {\bf Institutional review board (IRB) approvals or equivalent for research with human subjects}
    \item[] Question: Does the paper describe potential risks incurred by study participants, whether such risks were disclosed to the subjects, and whether Institutional Review Board (IRB) approvals (or an equivalent approval/review based on the requirements of your country or institution) were obtained?
    \item[] Answer: \answerNA{} 
    \item[] Justification: The paper does not involve crowdsourcing nor research with human subjects.
    \item[] Guidelines:
    \begin{itemize}
        \item The answer \answerNA{} means that the paper does not involve crowdsourcing nor research with human subjects.
        \item Depending on the country in which research is conducted, IRB approval (or equivalent) may be required for any human subjects research. If you obtained IRB approval, you should clearly state this in the paper. 
        \item We recognize that the procedures for this may vary significantly between institutions and locations, and we expect authors to adhere to the NeurIPS Code of Ethics and the guidelines for their institution. 
        \item For initial submissions, do not include any information that would break anonymity (if applicable), such as the institution conducting the review.
    \end{itemize}

\item {\bf Declaration of LLM usage}
    \item[] Question: Does the paper describe the usage of LLMs if it is an important, original, or non-standard component of the core methods in this research? Note that if the LLM is used only for writing, editing, or formatting purposes and does \emph{not} impact the core methodology, scientific rigor, or originality of the research, declaration is not required.
    \item[] Answer: \answerNA{} 
    \item[] Justification: The core method development in this research does not involve LLMs as any important, original, or non-standard components.
    \item[] Guidelines:
    \begin{itemize}
        \item The answer \answerNA{} means that the core method development in this research does not involve LLMs as any important, original, or non-standard components.
        \item Please refer to our LLM policy in the NeurIPS handbook for what should or should not be described.
    \end{itemize}

\end{enumerate}